# Learning social norms enhances compatibility in dynamic human–AI coordination

Yi Yang[1, 6], Siyuan Liu[1, 6], Xin Gao[1], Huamu Sun[1], Chao Liu[3, 4, 5], Qing Zhou[1], Bingbing Nie[1, 2, *]

## Abstract

Humans continuously coordinate with others in dynamic interactions, often through implicit, hard-to-quantify social norms that act as shared tacit expectations among interacting agents[1,2]. As AI agents, including large language models (LLMs), become embedded in daily life, they increasingly participate in such interactions and reshape social interaction structures[3,4]. Yet they often fail to coordinate with humans in an effective, considerate, and natural manner[5-8]. We hypothesize that this gap arises because existing approaches align model behavior with human demonstrations without explicitly quantifying the underlying norms that generate such behavior. We selected pedestrian-vehicle interaction as a representative dynamic interaction and developed a simplified experimental platform that captures its key interactive features. From 3,456 dynamic human interactions collected via this platform, we identified three principles underlying human social norms: outcome predictability, value alignment, and advantage awareness. Incorporating these principles into AI agents significantly improves human–AI coordination. In the closed-loop interaction task with humans, the social-norm-informed LLM achieved a nearly fourfold higher total score than the baseline strategy and outperformed human–human interactions by 43%. These findings indicate that formalizing tacit social norms into explicit, quantifiable principles can enable AI agents to achieve mutually beneficial coordination in dynamic interactions, supporting their more natural integration into human society.

## 1 Introduction

Dynamic interactions between humans and artificial intelligence (AI) are increasingly embedded in real-world physical and social environments[4,9,10]. In such settings, humans and AI agents engage in continuous, history-dependent, and mutually adaptive decision-making processes, where outcomes emerge through ongoing interactions between agents[11]. Representative scenarios include autonomous vehicles negotiating right-of-way with pedestrians[12], service robots navigating dense crowds[13], and AI agents engaging with humans in decision-making tasks such as logistics coordination and emergency response[14]. While many aspects of these interactions are regulated by explicit operational constraints, such as traffic regulations and physical safety limits[15-17], the constraints alone are insufficient to fully define successful coordination. A key limitation is that core processes of human coordination, such as anticipation, reciprocity, and mutual adaptation, are governed by implicit social norms[1,2]. If AI agents interacting with humans lack such norm-aware reasoning, inappropriate or even absurd consequences could follow. In terms of interaction quality, violations of social norms by AI can erode trust, increase cognitive friction, and induce frustration in human users[18-21]. In terms of physical safety, they introduce unpredictability into human–AI encounters that may escalate into hazardous outcomes[22]. For instance, pedestrians interacting with autonomous vehicles that lack social norms may experience heightened uncertainty and adopt defensive or assertive maneuvers, ultimately turning routine crossings into social friction and safety hazards[23] (Fig. 1a). Even large language models (LLMs), the dominant form of AI agents in today's human–AI interactions, frequently deviate from human social norms[24]. These issues suggest that the core challenge extends beyond rule compliance itself and lies in how to understand and adapt to the implicit social norms of humans.

[1]School of Vehicle and Mobility, Tsinghua University, Beijing, China. [2]State Key Laboratory of Intelligent Green Vehicle and Mobility, Tsinghua University, Beijing 100084, China. [3]State Key Laboratory of Cognitive Neuroscience and Learning & IDG/McGovern Institute for Brain Research, Beijing Normal University, Beijing, China. [4]Beijing Key Laboratory of Safe AI and Superalignment, Beijing, China. [5]Beijing Institute of AI Safety and Governance, Beijing, China. [6]These authors contributed equally: Yi Yang and Siyuan Liu. [*]Correspondence to: nbb@tsinghua.edu.cn

Existing efforts to endow AI with human social norms have almost exclusively focused on static or turn-based interactions, where agents deliberate before acting and receive feedback after each round[24-27]. By contrast, dynamic interactions unfold on fine-grained timescales, evolve through continuously coupled state trajectories, and tightly link one agent's instantaneous action to the other's preceding behavior[28,29]. These characteristics make social norms derived from static settings ill-suited for guiding dynamic human–AI coordination. A few studies have begun to address dynamic interaction, drawing on multi-agent reinforcement learning, imitation learning, inverse reinforcement learning, and rule-based planning[15,30-32]. However, these approaches typically formulate coordination as an explicit objective optimization problem, assuming that improving task-level objectives will naturally lead to socially compatible behavior. Empirical evidence suggests otherwise: reward optimization can induce non-human strategies[33], imitation learning fails to recover latent coordination structure[34], and rule-based constraints lack flexibility in dynamic environments[15]. We argue that existing methods fall short in dynamic interactions because they fail to explicitly quantify the social norms underlying human coordination. These norms cannot be fully revealed from demonstrations or optimized reward functions alone[34], but may instead be captured as a small set of quantifiable, transferable principles governing interaction dynamics.

Quantifying the social norms embedded in human dynamic interaction faces several key challenges (Fig. 1b). At the data level, social norms are difficult to extract because they are entangled with measurement errors and behavioral variations in noisy interaction data[35-37]. At the representation level, social norms are implicit and context-dependent, making them difficult to encode as explicit reward functions or optimization objectives[35,38]. At the decision-making level, effective human coordination does not arise from strict utility maximization. Rather, the normative structure underlying such coordination is reflected in intuitive, boundedly rational behaviors that may appear suboptimal under conventional optimization criteria[39].

To address these challenges, we integrate data collection, norm representation, and decision-making into a unified framework (Fig. 1c). We first established a simplified pedestrian-vehicle interaction task that preserves the core mechanics of dynamic coordination while minimizing irrelevant confounding factors. This setting yields clean human trajectories from which norm-related signals can be reliably isolated. Using this platform, we collected 3,456 dynamic interactions from 96 participants, enabling an open-loop evaluation of state-of-the-art LLMs through interactions with replayed human trajectories. We found that these models struggled to coordinate effectively with humans, leading to higher collision risks. Moreover, existing alignment strategies for static interactions failed to generalize to this dynamic setting, suggesting that dynamic coordination is governed by fundamentally different principles. Using the collected human interactions, we next inferred social norms by comparing successful and unsuccessful ones, rather than treating coordination as a purely optimization-based problem. This comparison revealed three principles underlying effective coordination: outcome predictability, value alignment, and advantage awareness. Together, these principles capture distinct dimensions of the normative structure and can each be operationalized quantitatively. Finally, we formalized these principles as explicit constraints and incorporated them into LLMs. Across multiple models, they improved coordination performance by reducing collision risk while preserving efficient passage behavior. As a result, AI agents moved beyond utility-driven decision-making toward socially normative behavior, ultimately outperforming humans.

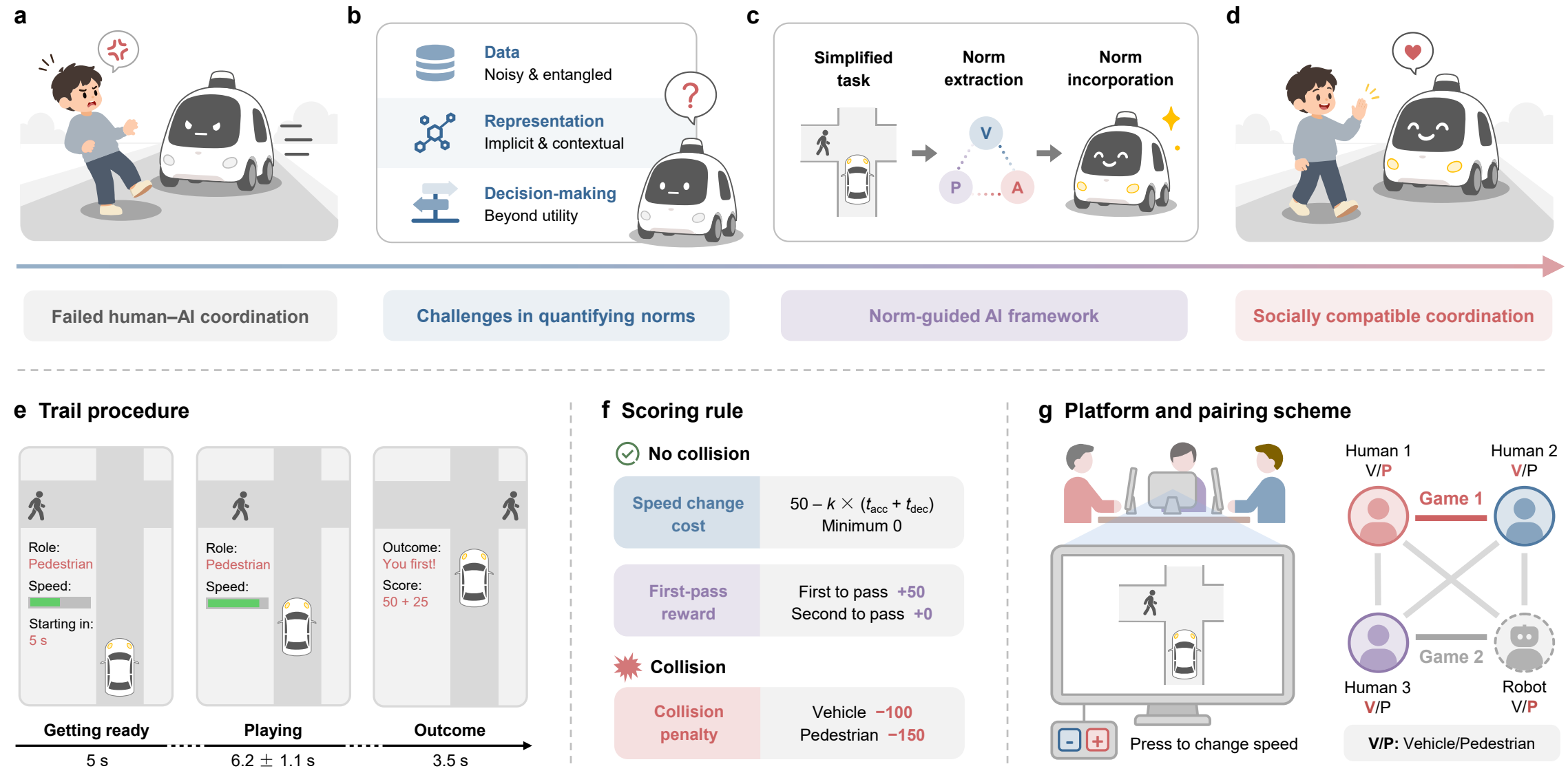

**Fig. 1 | a–d, Motivation and overview of the proposed framework. e–g, Pedestrian-vehicle dynamic game task and human data collection. a,** AI agents without social norms fail to coordinate effectively with humans. **b,** Challenges in quantifying implicit social norms across the data, representation, and decision-making levels. **c,** Overview of the proposed norm-guided AI framework (P, outcome predictability; V, value alignment; A, advantage awareness). **d,** Socially compatible coordination enabled by learned social norms. **e**, Trial procedure (pedestrian perspective). Before each trial, players are informed of their roles (pedestrian or vehicle) and initial speeds (indicated by a green progress bar). After a countdown, they adjust their speeds in real time until the trial ends, followed by outcome and score feedback. **f,** Scoring rule. Each player starts with 50 points, which decrease with the duration of speed adjustments (acceleration time, $t_{\mathrm{acc}}$; deceleration time, $t_{\mathrm{dec}}$). The first player to pass receives a 50-point bonus. A collision resets both scores to zero and incurs penalties of 100 and 150 points for the vehicle and pedestrian players, respectively. **g,** Left, experimental platform. Participants viewed the scenario and adjusted their speeds using the + and − keys. Right, pairing scheme. In each trial, two of the three participants were paired to interact, whereas the third interacted with a rule-based agent, allowing two independent games to run simultaneously. Participants were unaware of the identity of their interaction partner throughout the experiment.

Importantly, these gains extended beyond prompting. After distillation into lightweight policies, the three principles continued to guide effective coordination in real-time human–AI interaction: the social-norm-informed model achieved a total score nearly four times that of the baseline strategy, while also outperforming the human–human interaction reference by 43%. More broadly, these findings suggest that socially compatible AI in dynamic environments requires more than behavior imitation or objective optimization. It depends on explicitly representing the normative structure that governs human coordination. By extracting and quantifying these norms from human interaction data, we provide a framework for learning, transferring, and deploying socially compatible behaviors to enable safe and effective real-time human–AI coordination (Fig. 1d).

# 2 Results

## 2.1 Pedestrian-vehicle dynamic game

We focus on pedestrian-vehicle interactions, a representative form of dynamic interaction in everyday life. To collect human interaction data, we developed a simplified experimental platform that preserves only the essential characteristics of dynamic interactions, including continuous mutual adaptation, history-dependent decision-making, and tightly coupled behaviors (Fig. 1e–g). In each trial, one participant controls a pedestrian and the other controls a vehicle, each attempting to cross the intersection first without collision (Fig. 1e). They are able to adjust the speed of the controlled agent in real time using keyboard inputs. The scoring rule rewards coordination rather than pure competition or unconditional yielding: speed changes incur costs, passing the intersection first earns rewards, and collisions lead to penalties (Fig. 1f, Methods). Maximizing payoff therefore required players to anticipate each other and tacitly coordinate an appropriate passing order. To mitigate learning effects arising from repeated games[24,40], we designed a pairing scheme (Fig. 1g, Methods). Specifically, each experimental session involved three participants, with two paired to interact in each trial and the third interacting with a rule-based virtual agent (see Supplementary Note 1.5). Throughout the experiment, participants were unaware of the identity of their partner. The experiment generated a dataset of 3,456 human–human interactions from 96 participants, used for extracting social norms of successful coordination, as well as establishing a human baseline for quantitative evaluation of LLM behavior.

## 2.2 LLMs struggle to coordinate with humans

LLMs, as the current mainstream AI agents, were evaluated in the pedestrian-vehicle interaction task. Despite LLM inference latency[41], we developed a keyframe-based open-loop evaluation paradigm that closely approximates closed-loop interaction (Methods). We extracted 6,924 keyframes reflecting clear decision intent from the human interaction data. At each keyframe, the LLM replaced the human player and generated its next actions based on the opponent's recorded trajectory. We evaluated LLM coordination using two representative metrics: mean collision probability $\bar{p}_{\mathrm{C}}$ and mean first-passage probability $\bar{p}_{\mathrm{F}}$. Here, $\bar{p}_{\mathrm{C}}$ measures interaction safety (lower is better), whereas $\bar{p}_{\mathrm{F}}$ measures assertive passage efficiency (higher is better), both estimated by a long short-term memory (LSTM) network trained on human interaction data (Methods).

The five selected LLMs (i.e., DeepSeek-V4-Flash, Gemini-3.1-Flash-Lite, Doubao-Seed-2.0-Lite, Llama-4-Maverick, and GPT-5.4-Mini) generally struggled to achieve coordination with humans (Fig. 2, Base). In the Base condition, the prompt provided only an objective description of the interaction task and scenario, without normative or subjective guidance (see Supplementary Note 3). Compared with human behavior, DeepSeek, Gemini, and Llama significantly increased $\bar{p}_{\mathrm{C}}$ ($P < 0.001$) without a corresponding increase in $\bar{p}_{\mathrm{F}}$. Doubao increased both $\bar{p}_{\mathrm{C}}$ and $\bar{p}_{\mathrm{F}}$ ($P < 0.001$), indicating improved passage efficiency at the expense of safety. Although GPT significantly reduced $\bar{p}_{\mathrm{C}}$ ($P < 0.001$), it did so at the cost of a significant decrease in $\bar{p}_{\mathrm{F}}$ ($P < 0.001$). These results indicate that LLMs are unable to spontaneously achieve effective coordination with humans in dynamic interactions.

Mainstream alignment approaches shown to be effective in static games failed to improve LLM coordination in dynamic interactions. Building on the Base prompt, we first evaluated the Social Chain-of-Thought (SCoT) prompt, which instructed the LLM to infer the opponent's intent before acting accordingly[24] (see Supplementary Note 3.3). Across multiple LLMs, $\bar{p}_{\mathrm{C}}$ remained above the human baseline, with no significant improvement over Base (Fig. 2, SCoT). To further reduce LLM collision risk, we used a risk-aversion (RA) prompt that prioritized collision avoidance over being the first to pass[25] (Supplementary Note 3.3). Although RA effectively suppressed collisions, it did so at the cost of losing assertiveness. $\bar{p}_{\mathrm{F}}$ was significantly lower than the human baseline ($P < 0.001$), reflecting overly cautious behavior (Fig. 2, RA). In summary, neither intervention achieved an appropriate balance between safety and assertiveness. SCoT failed to mitigate aggressive behavior, whereas RA achieved safety at the cost of excessive caution. These results further suggest that deliberative reasoning strategies developed for static games is insufficient to internalize the logic of human coordination in dynamic interactions.

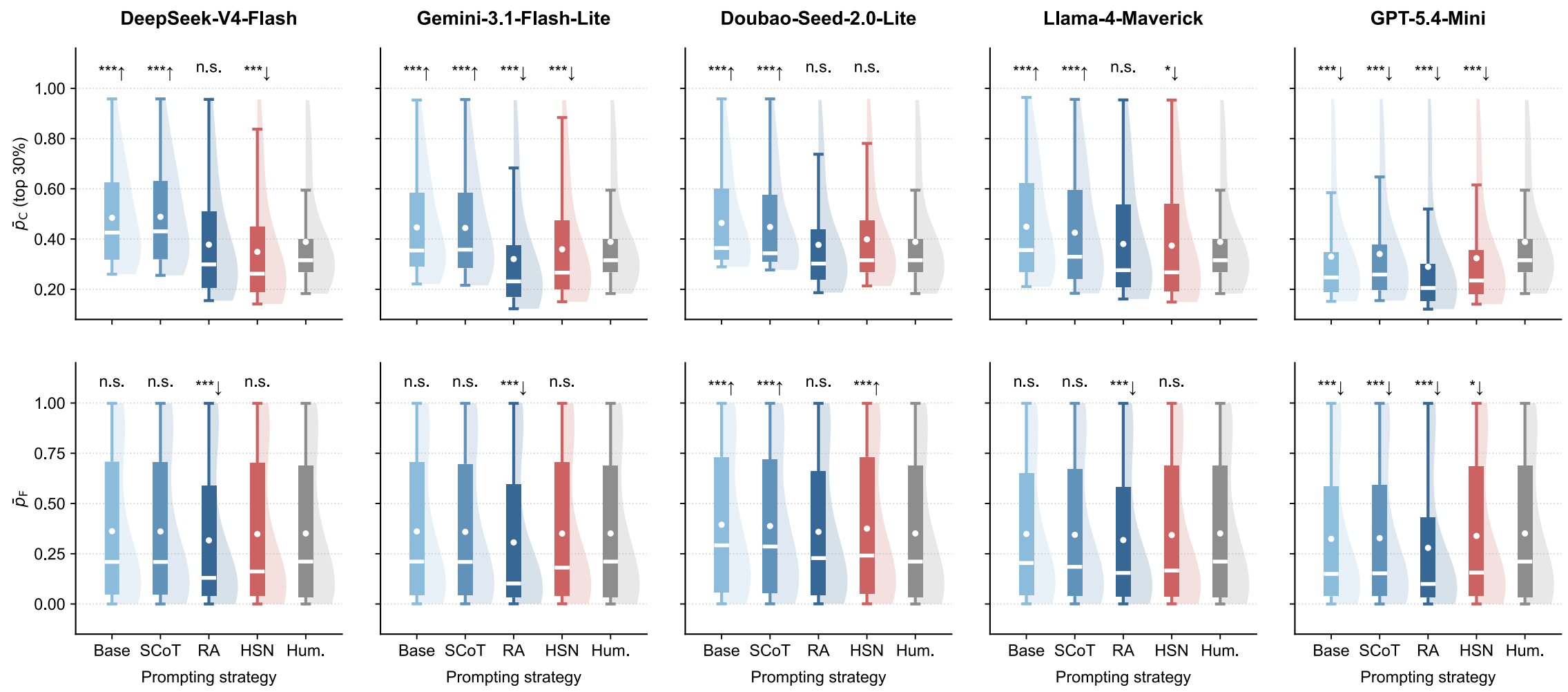

**Fig. 2 | Performance of five LLMs under different prompting conditions in open-loop evaluation.** HSN, human social norm; SCoT, Social Chain-of-Thought; RA, risk aversion; Hum., human baseline; $\bar{p}_{\mathrm{C}}$, mean collision probability; $\bar{p}_{\mathrm{F}}$, mean first-passage probability. Each prompting condition was evaluated on the same set of 6,924 keyframes. Because most $\bar{p}_{\mathrm{C}}$ values clustered near zero, only the upper 30% of the distribution is shown for clarity, whereas the full distribution of $\bar{p}_{\mathrm{F}}$ is shown. Semi-violin plots show data distributions. Boxes indicate the interquartile range (IQR), center lines the median, whiskers extend to 1.5 × IQR, and white dots indicate the means. Asterisks denote statistical significance relative to the human baseline based on two-sided permutation tests (10,000 resamples): * $P < 0.05$, ** $P < 0.01$, *** $P < 0.001$; n.s., not significant. Upward and downward arrows indicate significantly higher or lower values than the human baseline, respectively.

## 2.3 Principles underlie human social norms

In dynamic interactions, three coupled principles distinguish successful coordination from failed coordination: outcome predictability, value alignment, and advantage awareness. These principles capture distinct yet complementary aspects of human coordination, and can each be quantified using a dedicated metric (Methods). To identify these principles, we categorized 3,456 human–human interactions according to the combined score of both players, which served as an overall measure of interaction quality (Fig. 3a). For non-collision trials, the theoretical range of total scores was divided into three equal intervals. Each interval was subsequently split according to whether the vehicle or the pedestrian crossed first, yielding six categories (C1–C6). All collision trials were assigned the category C7. In C1 and C2, smooth and efficient coordination emerges when one player yields at a constant speed while the other accelerates through. In C3 and C4, both players initially accelerate, after which one decelerates to yield. In C5 and C6, both players first accelerate and then decelerate before one passes. In C7, neither player yields, leading to a collision.

**Outcome predictability.** Effective coordination requires that the interaction remain legible to both parties, enabling each to anticipate how it is likely to unfold. When intentions are ambiguous or behavior is inconsistent, players tend to hesitate or develop misaligned expectations, increasing the risk of coordination failure[22]. By contrast, in high-quality coordination, the interaction outcome becomes apparent early and remains predictable over time. We quantified this property using game entropy, defined as the information entropy of the real-time outcome distribution predicted by an LSTM (Methods). Low entropy indicates that one outcome dominates and the interaction is readily interpretable, whereas high entropy reflects persistent ambiguity. Game entropy showed distinct trajectories across interaction categories (Fig. 3b, left): it decayed rapidly in C1–C2, declined during the middle phase in C3–C6, and remained high until the late stage in C7. We further computed cumulative entropy (Methods) and identified a quantitative threshold separating high-quality coordination from less successful coordination (Fig. 3b, right).

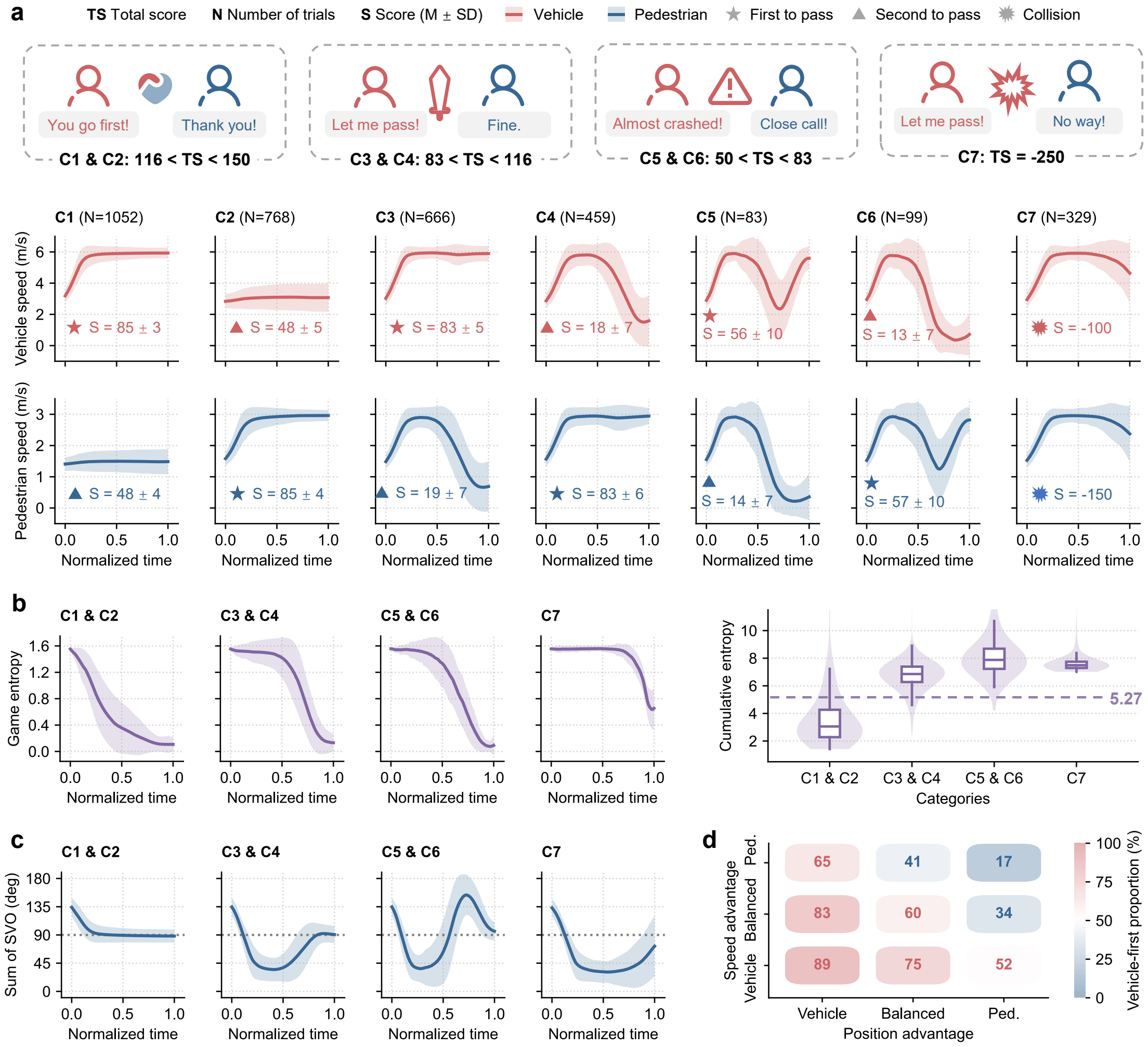

**Fig. 3 | Multi-dimensional analysis of human interaction data across coordination quality categories.** Trial curves within each category were time-normalized and averaged (mean ± s.d.). **a,** Top, classification of human interactions according to the combined score of both players, with representative interaction patterns for each category. Bottom, vehicle and pedestrian speed profiles for each category. **b,** Left, game entropy over time across categories. Right, distribution of cumulative entropy (time integral of game entropy), shown as violin plots overlaid with box plots (center line, median; box, IQR; whiskers, 1.5 × IQR). The threshold (5.27), defined as the mean of the upper quartile of C1–C2 and the lower quartile of C3–C4, separates high-quality coordination from other interactions. **c,** Sum of SVO angles over time across categories. **d,** Proportion of vehicle-first trials under different initial position and speed conditions based on C1–C6 interactions (collision trials excluded). Both factors comprised three levels: vehicle-advantaged, balanced, and pedestrian-advantaged.

**Value alignment.** Successful coordination depends on complementary value orientations. For example, when a pedestrian and a vehicle meet at an intersection, coordination emerges not through explicit Nash-equilibrium reasoning but through the implicit alignment of social preferences over each other's outcomes[11]. High-quality coordination therefore requires one party to claim priority while the other yields, rather than both racing ahead or both holding back. We used social value orientation (SVO), a psychometric construct, to quantify the weight a player assigns to the other player's payoff relative to their own. At each timestep, we computed each player's SVO angle from the ongoing game state, where 0° denotes egocentrism and 90° altruism (Methods). The sum of the two players' SVO angles provides a direct measure of value alignment: it falls below 90° when both players race ahead, exceeds 90° when both yield, and approaches 90° only when their value orientations are complementary. This metric closely tracks interaction quality (Fig. 3c): it rapidly converges to 90° in C1–C2, dips into competition before realigning in C3–C4, oscillates without converging in C5–C6, and remains below 90° in C7.

**Advantage awareness.** In dynamic interactions, the two parties often differ in relative advantages, and effective coordination requires that these advantages be recognized[42-44]. Treating the interaction as a symmetric optimization problem can lead agents to misjudge what constitutes socially acceptable behavior[12]. In this study, relative advantage arises from both agent-intrinsic and situation-dependent asymmetries. Agent-intrinsic asymmetry reflects unequal collision penalties, with pedestrians penalized more heavily than vehicles. Situation-dependent asymmetry is determined by the initial positions and speeds, yielding three conditions: vehicle-advantaged, pedestrian-advantaged, and balanced (see Supplementary Note 1.3). Effective coordination emerges when both forms of relative advantage are respected: the advantaged player tends to maintain priority, whereas the other yields. As shown in Fig. 3d, statistics from C1–C6 reveal that the passing order shifts with the objective advantage determined by the initial positions and speeds. Meanwhile, agent-intrinsic asymmetry introduces a systematic bias: even under identical initial conditions, the vehicle is more likely to pass first. In C7, failure to respect this advantage leads to collisions.

## 2.4 Social norms enable human–AI coordination

We formalized the three principles as explicit representations of human social norms and incorporated them into LLM prompts. Across all five models, the human social norms (HSN) prompting strategy consistently improved coordination with humans in the open-loop evaluation, outperforming both the baseline prompt (Base) and existing static-interaction strategies (SCoT and RA). As shown in Fig. 2, HSN achieved significantly lower $\bar{p}_{\mathrm{C}}$ than the human baseline for DeepSeek ($P < 0.001$), Gemini ($P < 0.001$), and Llama ($P < 0.05$), while maintaining $\bar{p}_{\mathrm{F}}$ at the human level. For Doubao, HSN significantly increased $\bar{p}_{\mathrm{F}}$ ($P < 0.001$) without increasing $\bar{p}_{\mathrm{C}}$. For GPT, although all prompting strategies significantly reduced $\bar{p}_{\mathrm{C}}$ relative to the human baseline ($P < 0.001$), HSN produced the smallest reduction in $\bar{p}_{\mathrm{F}}$.

HSN generalized beyond interactions with replayed trajectories to real-time closed-loop coordination with human participants. To enable real-time deployment, we distilled the decision-making behavior of DeepSeek-V4-Flash under four prompting strategies into lightweight surrogate policies (Methods), which were evaluated on the original experimental platform (Methods). Fig. 4a shows the mean self-score and mean total score for each strategy in human–AI interactions, together with the population-level human–human performance computed from the original interaction dataset (Hum.). Compared with the Base strategy, HSN achieved approximately 2.5-fold higher mean self-score and nearly 4-fold higher mean total score. Relative to RA, SCoT, and Hum., it increased mean total score by 12%, 51%, and 43%, respectively, and mean self-score by 11%, 9%, and 24%, respectively. Notably, human participants paired with HSN achieved the highest mean self-score (68) across all strategies, indicating that HSN improved not only the agent's performance but also that of its human partners, enabling mutually beneficial coordination.

To test whether social norms can be acquired from trajectories alone, we trained a set of imitation-learning agents on human–human interaction data. Four lightweight agents were trained on datasets spanning progressively broader interaction quality: H1–2 (trained on highest-quality interactions, C1–C2), H1–4 (trained on C1–C4), H1–6 (trained on C1–C6), and Hum. (trained on all interactions). Together with the LLM-distilled agents (Base, SCoT, RA, and HSN), these models were paired with each other to complete 1,000 interactions (Methods). We then computed the mean scores of each model (Fig. 4b) and strategy pair (Fig. 4c). HSN consistently outperformed all other prompting strategies, and none of the imitation-learning agents matched its performance. Notably, H1–2 even underperformed H1–4, H1–6 and Hum., suggesting that training exclusively on successful interactions limits generalization because the model never learns to recover from failed coordination. Together, these results show that the logic underlying human coordination cannot be recovered from trajectory imitation, underscoring the need to explicitly represent social norms.

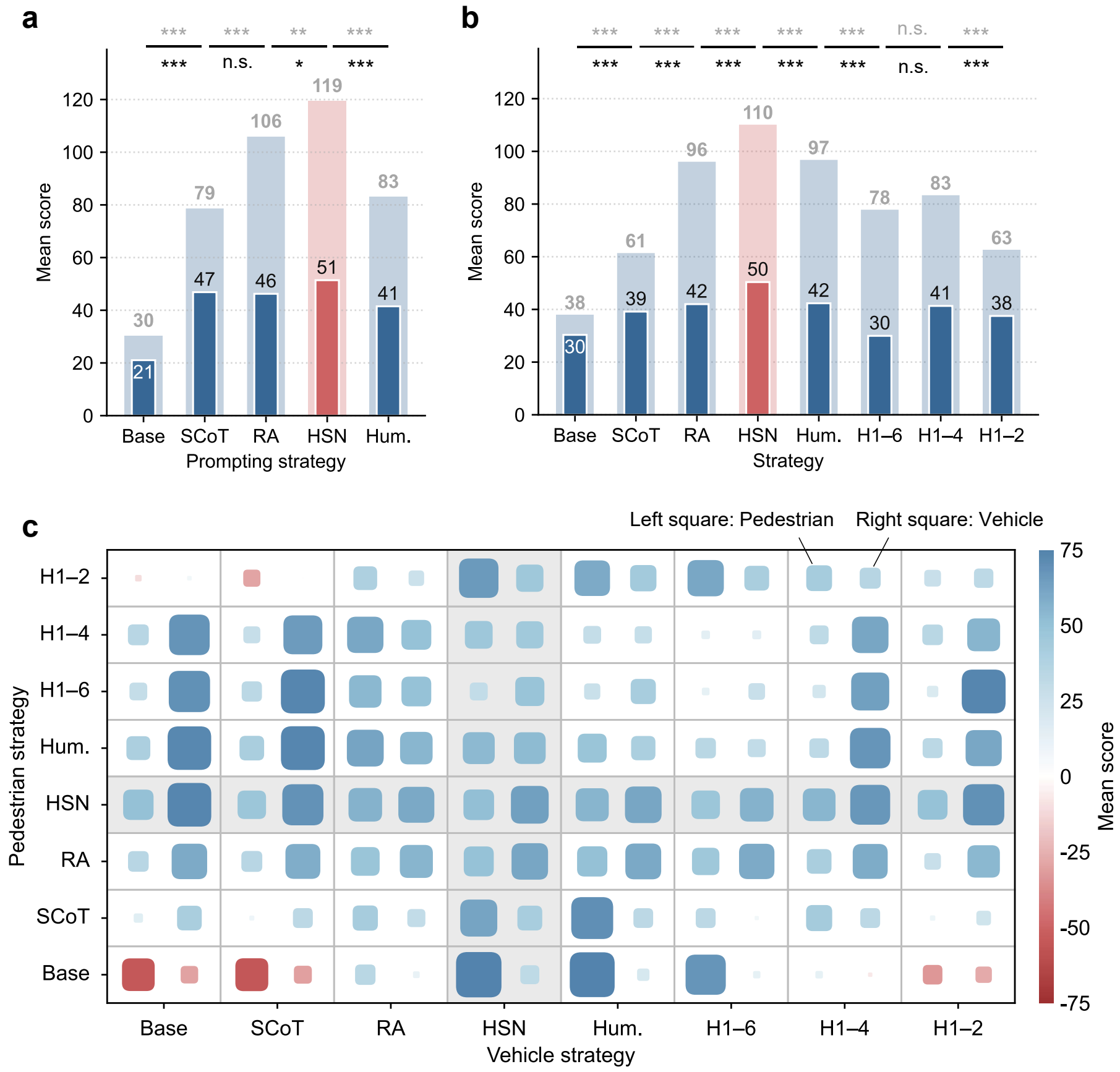

**Fig. 4 | Performance of different closed-loop interaction strategies. a,** Mean self-scores (dark bars; lower black asterisks) and mean total scores (light bars; upper gray asterisks) for distilled models under four prompting strategies in human–AI interactions, with human–human interactions (Hum.) shown as a reference. Asterisks indicate Welch's t-test for adjacent comparisons: * $P < 0.05$, ** $P < 0.01$, *** $P < 0.001$; n.s., not significant. **b,** Mean self-scores and mean total scores for different strategies in pairwise closed-loop games. Bar colors and significance notation follow the conventions in **a**. **c,** Mean joint-score matrix for pairwise closed-loop interactions. Each cell shows the pedestrian score (left square) and vehicle score (right square). Color indicates the mean score, and square size is proportional to its absolute value.

Ablation analysis demonstrated that all three principles are necessary. We encoded outcome predictability (P), value alignment (V), and advantage awareness (A) into LLM prompts individually and in combination, and evaluated each strategy in the open-loop setting (Fig. 5a). Adding any single principle (+A, +V, or +P) significantly reduced $\bar{p}_C$ ($P < 0.001$), with the lowest $\bar{p}_C$ achieved when all three principles were combined (HSN), while $\bar{p}_F$ remained unchanged. Notably, in the open-loop evaluation, advantage awareness contributed less to reducing $\bar{p}_C$ than the other two principles. In some cases, adding A did not further reduce $\bar{p}_C$ (e.g., +V → +AV and +P → +AP). By contrast, self-play revealed that advantage awareness is equally essential for stable coordination (Fig. 5b). When each distilled strategy interacted with an identical copy of itself, strategies lacking A consistently achieved lower scores, whereas incorporating A markedly improved performance. This is because only advantage-aware policies enable identical agents to adopt complementary roles, with one claiming priority and the other yielding. Together, these findings show that the three principles are complementary rather than redundant: each resolves a distinct coordination failure mode, and only their combination reconstructs the implicit social norms underlying human dynamic coordination.

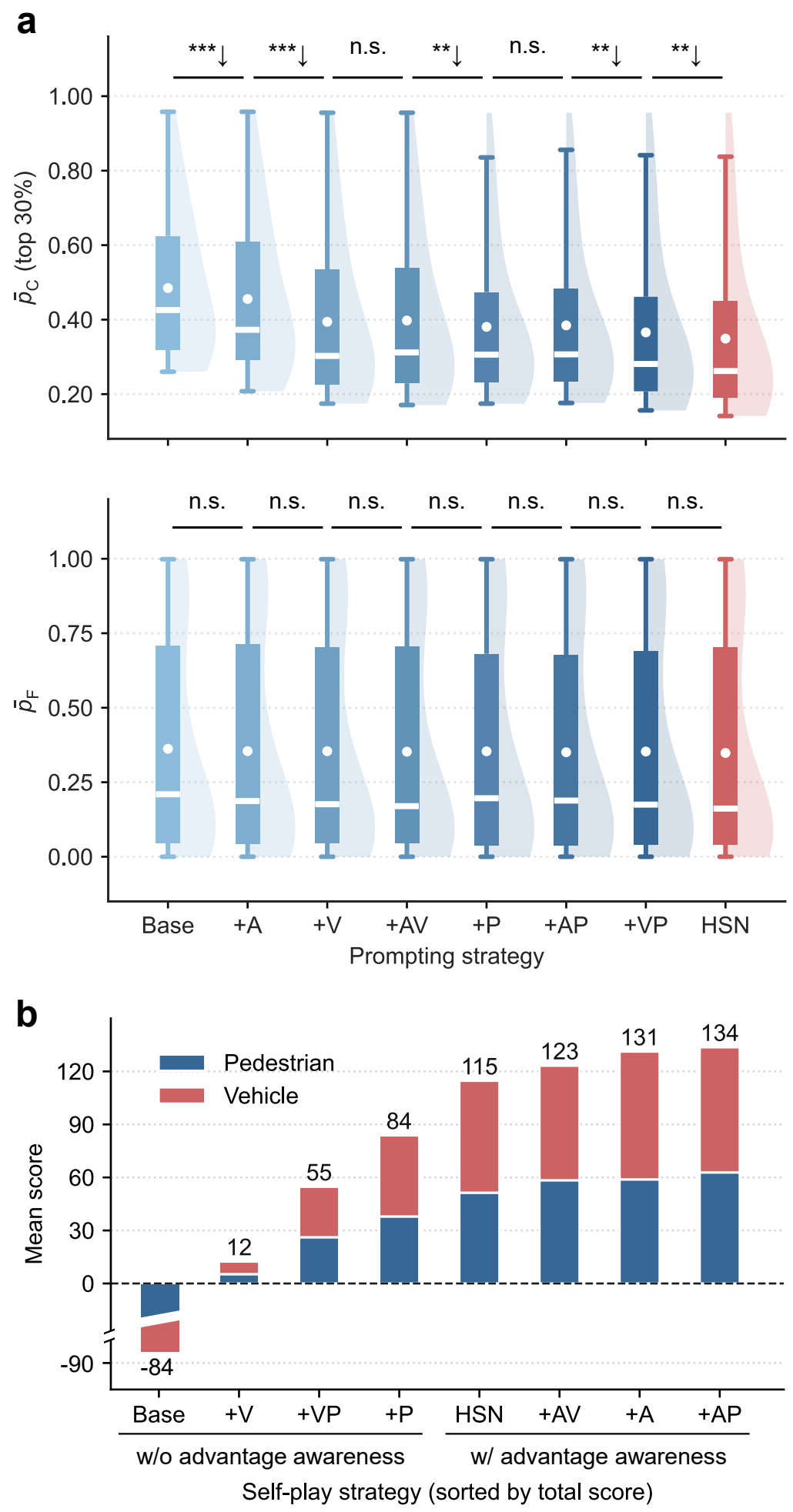

**Fig. 5 | Ablation analysis of the three principles** (P, outcome predictability; V, value alignment; A, advantage awareness). **a,** Performance of different principle combinations in the open-loop evaluation on DeepSeek-V4-Flash ($\bar{p}_C$, mean collision probability; $\bar{p}_F$, mean first-passage probability). Each prompting condition was evaluated on the same set of 6,924 keyframes. Because most $\bar{p}_C$ values clustered near zero, only the upper 30% of the distribution is shown for clarity, whereas the full distribution is shown for $\bar{p}_F$. Semi-violin plots show the distributions; boxes indicate the IQR, center lines the median, whiskers extend to 1.5 × IQR, and white dots denote the mean. Asterisks indicate statistical significance between adjacent conditions based on two-sided permutation tests (10,000 resamples): * $P < 0.05$, ** $P < 0.01$, *** $P < 0.001$; n.s., not significant. Arrows indicate that the condition on the right is significantly higher (↑) or lower (↓) than that on the left. **b,** Mean self-play scores for distilled models under different principle combinations, ordered by increasing total score. Red and blue bars represent vehicle and pedestrian scores, respectively, and the mean total score is shown above each bar. The Base bar is truncated.

# 3 Discussion

As AI agents become increasingly embedded in human society, enabling socially compatible human–AI interactions has become an urgent challenge. Existing approaches largely address this problem through output alignment, optimizing model behavior to better match human values[45,46]. Here, we argue that social compatibility requires more than behavioral alignment: it depends on identifying the implicit social norms that govern human interaction. For AI agents, translating these norms into computational rules remains challenging because they are embedded in noisy behavioral data, represented at an abstract social level, and expressed through boundedly rational decisions. By analyzing regularities in human behavior within a simplified pedestrian-vehicle dynamic interaction task, we show that implicit social norms can be recovered and formalized as three coupled quantitative principles: outcome predictability, value alignment, and advantage awareness. Explicitly encoding these principles into LLM prompts substantially improved safety while preserving efficient passage in closed-loop interactions with human participants, resulting in a nearly fourfold increase in total score over the baseline strategy and a 43% improvement over human–human interactions. These findings suggest that making implicit social norms explicit can not only reproduce human coordination but also surpass it.

The three principles are mutually reinforcing rather than redundant because each addresses a distinct challenge of dynamic coordination, and only their combination captures the implicit social norms underlying successful human interaction. Outcome predictability reduces uncertainty about how an interaction will unfold, enabling both parties to anticipate its likely outcome within a limited reaction window. Value alignment encourages interacting agents to make complementary rather than conflicting decisions over priority, thereby fostering compatible value orientations and coordinated behavior. Advantage awareness accounts for the asymmetries inherent in social interaction by conditioning decisions on relative advantages, enabling agents to adopt behaviors appropriate to their respective positions.

Our findings highlight a key property of social compatibility in dynamic interactions: successful coordination is not simply an objective optimization problem, but depends on balancing multiple normative principles. The poor performance of the Base strategy suggests that such a balance does not reliably emerge from pretraining alone[45]. Existing interventions likewise showed clear limitations: SCoT produced only marginal improvements, and RA achieved safety largely by sacrificing efficiency. By contrast, HSN simultaneously resolved multiple coordination failures, reducing collisions while preserving efficient passage. More fundamentally, these results suggest that the normative structure underlying successful coordination cannot be recovered solely from behavioral imitation or task rewards. Instead, what must be extracted is the transferable normative structure that gives rise to coordinated behavior, represented here by three principles identified from human interactions. Closed-loop evaluations showed that the benefits of these principles were preserved after distillation, whereas imitation learning from high-quality human trajectories failed to recover HSN-level performance. Similarly, the same principles generalized to reinforcement learning[47], remaining effective even when implemented solely as reward terms (Supplementary Note 7). In contrast, agents optimized using task rewards alone still performed poorly.

A limitation of this study is the inference latency of current LLMs, which precludes direct closed-loop interaction with humans[41]. We therefore adopted a two-stage evaluation framework combining open-loop evaluation with LLMs and closed-loop evaluation with distilled policies. As inference efficiency continues to improve, direct closed-loop evaluation of LLMs should become feasible. Another limitation is that the present study focuses on identifying transferable normative principles from observed human interactions rather than explaining their origins through a mechanistic theory. Developing forward computational models that derive these principles from human decision-making may provide a stronger mechanistic account of the social norms. Beyond pedestrian-vehicle coordination, the norm-extraction framework introduced here may generalize to a broad range of dynamic human–AI interactions, from service and delivery robots to AI agents collaborating with human operators. Despite their diversity, these scenarios share a fundamental structure of dynamic interaction.

As AI systems become increasingly autonomous, maintaining alignment with implicit human social norms will be essential for safe and acceptable deployment. Our findings suggest that the norms underlying coordination in dynamic human interactions are not entirely inaccessible. They can be characterized quantitatively, at least to the extent required to induce socially considerate AI behavior. Represented as computable principles of outcome predictability, value alignment, and advantage awareness, these norms improve coordination when incorporated into AI agents, enabling performance beyond the human baseline. More broadly, our findings suggest that achieving social compatibility requires not only optimizing task objectives or reproducing human behavior, but also explicitly representing the normative structure that governs human coordination. By extracting, formalizing, and transferring implicit social norms from human behavior, this work provides a path toward socially compatible AI agents that can integrate into human society more naturally and effectively.

# 4 Methods

## 4.1 Human–human game data collection

**Participants.** A total of 96 adults were recruited, comprising 24 males under 30 years of age (mean age ± s.d., 24.5 ± 3.6 years), 24 females under 30 (22.9 ± 3.1 years), 24 males over 30 (39.6 ± 6.5 years), and 24 females over 30 (41.9 ± 7.6 years). Participant details are provided in Supplementary Note 2.1. Each age-gender cohort was further divided into eight groups of three participants. Each group completed one experimental session together with a rule-based virtual agent, with participant pairings assigned as described in Design and pairing scheme. No participants were excluded from the analysis. All participants provided informed consent, and the study was approved by the Institutional Review Board of Tsinghua University. Participants were compensated according to their final ranking based on cumulative scores across the experiment, receiving between ¥75 and ¥150, with payments for intermediate rankings determined by linear interpolation. To examine whether demographic factors influenced coordination performance, we conducted a two-way ANOVA on mean individual scores. Neither age ($F(1, 92) = 3.90$, $P = 0.051$), gender ($F(1, 92) = 2.00$, $P = 0.161$), nor their interaction ($F(1, 92) = 2.22$, $P = 0.140$) showed a significant effect, indicating that the observed coordination patterns were not systematically associated with participant demographics.

**Task and trial procedure.** In the pedestrian-vehicle dynamic interaction task, two participants controlled a pedestrian and a vehicle, respectively, and attempted to traverse the intersection before the other (Fig. 1e). As shown in Fig. 1f, each trial began with both players receiving 50 initial points. During the interaction, players could adjust their speed in real time at the cost of consuming these points. The game terminated when either player had completely crossed the intersection or when a collision occurred. If no collision occurred, the player who crossed first received an additional 50-point reward. In the event of a collision, both players' scores were reset to zero, and additional penalties of 100 and 150 points were imposed on the vehicle and pedestrian players, respectively. Detailed reward rules and scenario parameters are provided in Supplementary Note 1.

As shown in Fig. 1e, each trial began with a 5-s countdown during which participants were informed of their assigned role and initial speed. During the trial, participants could adjust the speed of their assigned agent using the + and − keys. Trials lasted 6.2 ± 1.1 s (mean ± s.d.) on average, during which the trajectories and control inputs of both agents were recorded at 20 Hz. At the end of each trial, participants received feedback on the outcome and their score, which was displayed for 3.5 s. In non-collision trials, the score was presented as the sum of the remaining points and the priority reward. In collision trials, the corresponding penalty (−100 or −150 points) was displayed directly.

**Design and pairing scheme.** To mitigate strategy learning and reciprocity arising from repeated games, thereby preserving the independence of individual trials, we employed a predefined pairing scheme (Fig. 1g). Each experimental session comprised three human participants and one rule-based agent. In each trial, two participants were paired with one another, whereas the third interacted with the agent, allowing two games to proceed simultaneously. Participants remained unaware of the identity of their interaction partner throughout the experiment. The rule-based agent implemented three behavioral strategies (yielding, neutral, and aggressive; see Supplementary Note 1.5), each presented with equal frequency to expose participants to diverse interaction styles. Post-experiment debriefing confirmed that participants could not reliably distinguish virtual from human opponents. Moreover, mean participant scores did not differ significantly between interactions with human and virtual opponents ($P > 0.05$).

Five experimental factors were systematically varied across trials: participant pairing (three levels: corresponding to the three possible pairings), role (pedestrian or vehicle), initial position (balanced, vehicle-advantaged, or pedestrian-advantaged), initial speed (balanced, vehicle-advantaged, or pedestrian-advantaged), and scenario orientation (horizontal or vertical). Chi-square tests confirmed that both initial position ($\chi^2(4) = 444.26$, $P < 0.001$, Cramér's $V = 0.254$) and initial speed ($\chi^2(4) = 217.16$, $P < 0.001$, $V = 0.177$) significantly influenced outcomes, confirming the effectiveness of the intended asymmetry manipulation. Scenario orientation also showed a statistically significant but negligible effect ($\chi^2(2) = 15.50$, $P < 0.001$, $V = 0.067$), possibly reflecting minor perceptual asymmetries. Because orientation was fully counterbalanced across the experiment, it was not considered further. The full factorial combination of these factors generated 108 unique experimental conditions (3 × 2 × 3 × 3 × 2), each presented once, resulting in 108 trials completed in six blocks separated by 1-min breaks. Before the formal experiment, participants completed practice trials to familiarize themselves with the task and scoring rules. Experimental sessions lasted approximately 45 min. In total, 3,456 human–human interactions were collected.

## 4.2 Open-loop LLM coordination evaluation

Current LLMs exhibit substantial inference latency, precluding real-time closed-loop interaction with humans. We therefore developed a short-term open-loop evaluation paradigm in which LLMs interact with recorded human trajectories at selected keyframes. We defined three types of keyframes: acceleration frames, corresponding to transitions from constant-speed to accelerating motion; deceleration frames, corresponding to transitions from constant-speed to decelerating motion; and constant-speed frames, randomly sampled from constant-speed segments longer than 20 frames. Each keyframe represents an explicit human decision intended to influence the course of the interaction. We sampled the three keyframe types in a 1:1:1 ratio, resulting in a total of 6,924 keyframes extracted from the human interaction dataset.

At each keyframe, the LLM assumed control of the corresponding agent. During the subsequent 2-s evaluation window, the LLM received prompts and generated decisions (target speeds) at 0.5-s intervals. The prompts provided a structured description of the task, reward rules, current scenario state, and other relevant information (see Supplementary Note 3). The opposing agent followed the original human trajectory throughout the evaluation. Because the replayed human trajectory did not respond to the LLM's actions, extending the evaluation window would gradually reduce the realism of the interaction. We therefore limited the evaluation window to 2 s. The underlying physical simulation was updated and recorded at 20 Hz. This framework enabled evaluation of LLM decision-making in response to human behavior under controlled open-loop conditions.

We evaluated LLM coordination performance using two metrics: the mean collision probability $\bar{p}_{\mathrm{C}}$ and the mean first-passage probability $\bar{p}_{\mathrm{F}}$ over the 2-s window following LLM intervention. Both metrics were estimated using an LSTM network trained on human interaction data (see Supplementary Note 6). At each timestep $t$, the model received an 8-dimensional kinematic feature vector $x_t \in \mathbb{R}^8$, consisting of the positions, speeds, distances to the conflict point, and time-to-conflict-point for both agents. All continuous features were normalized. The input sequence $X = [x_1, \ldots, x_{20}]$ spanned the preceding 1 s (20 frames). A two-layer LSTM encoded the sequence into a latent representation $h_{20}^{(2)} = \mathrm{LSTM}(X)$, which was passed to a multilayer perceptron (MLP) classification head to produce logits $z \in \mathbb{R}^3$. A softmax layer then generated the probability distribution $\mathbf{p} = (p_{\mathrm{P}}, p_{\mathrm{V}}, p_{\mathrm{C}})$ over the three possible outcomes (pedestrian first, vehicle first, or collision), where $p_i = \frac{\exp(z_i)}{\sum_{j=1}^{3} \exp(z_j)}, i \in \{\mathrm{P}, \mathrm{V}, \mathrm{C}\}$. The network was trained using cross-entropy loss. During inference, probabilities were estimated using a temperature-scaled softmax, $\tilde{p}_i = \frac{\exp(z_i/T)}{\sum_{j=1}^{3}\exp(z_j/T)}$, with a fixed temperature of $T = 4$ to reduce short-term probability fluctuations. Let $t_k$ denote the timestep corresponding to the keyframe. The mean collision probability was computed as $\bar{p}_\mathrm{C} = \frac{1}{40}\sum_{t=t_k+1}^{t_k+40}\tilde{p}_\mathrm{C}^{(t)}$. The mean first-passage probability was computed as $\bar{p}_\mathrm{F} = \frac{1}{40}\sum_{t=t_k+1}^{t_k+40}\tilde{p}_\mathrm{P}^{(t)}$ when the LLM controlled the pedestrian, and $\bar{p}_\mathrm{F} = \frac{1}{40}\sum_{t=t_k+1}^{t_k+40}\tilde{p}_\mathrm{V}^{(t)}$ when it controlled the vehicle.

In the dynamic interaction task, decisions rely on limited historical information and must be made under time constraints. Accordingly, we selected LLMs based on two criteria: sufficient reasoning capability to interpret dynamic social cues, and lightweight design to simulate the time and cognitive resource constraints faced by humans in dynamic interaction. Following these criteria, we evaluated five representative LLMs: DeepSeek-V4-Flash, Gemini-3.1-Flash-Lite, Doubao-Seed-2.0-Lite, Llama-4-Maverick, and GPT-5.4-Mini. These models represent the latest versions available to us. For each model, we tested four prompting strategies: Base, SCoT, RA, and HSN (see Supplementary Note 3 for detailed prompts). To assess the generalizability of our findings, the open-loop evaluation was repeated across all five models. We further analyzed their detailed behavioral responses to characterize how different prompting strategies shaped model decision-making across model families (see Supplementary Note 4).

## 4.3 Quantification of the three principles

**Game entropy.** We quantified the uncertainty of the interaction at timestep $t$ using Shannon entropy:

$$h^{(t)} = -\sum_{i\in\{\mathrm{P,V,C}\}} p_i^{(t)} \log_2 p_i^{(t)} \tag{1}$$

where $p_i^{(t)}$ $(i \in \{\mathrm{P}, \mathrm{V}, \mathrm{C}\})$ denotes the predicted probability of each outcome at timestep $t$, estimated by the outcome prediction network. Lower values of $h^{(t)}$ indicate that one outcome dominates the probability distribution and the interaction outcome is more predictable, whereas higher values indicate greater uncertainty due to a more uniform distribution over the three outcomes. The cumulative game entropy was defined as the temporal integral of game entropy, $H_\mathrm{cum} = \int_0^T h^{(t)}\,\mathrm{d}t$.

**Social Value Orientation (SVO).** We quantified SVO in real time as an angle $\theta^{(t)}$. At timestep $t$, each player's reward to self $r_s^{(t)}$ and reward to other $r_o^{(t)}$ were computed from the kinematic states of the two agents:

$$\begin{aligned} r_\mathrm{s}^{(t)} &= \exp\left(\alpha\frac{s_\mathrm{s}^{(t)}}{t}\right)\cdot\exp\left(\alpha v_\mathrm{s}^{(t)} - v_\mathrm{mean}\right) \\ r_\mathrm{o}^{(t)} &= \exp\left(\alpha\frac{s_\mathrm{o}^{(t)}}{t}\right) \end{aligned} \tag{2}$$

where $s^{(t)}$ denotes the distance traveled, and $s^{(t)}/t$ is the historical average speed, reflecting the overall passage efficiency of each agent. The scaling coefficient $\alpha$ (pedestrian: 2; vehicle: 1) compensates for differences in speed ranges between pedestrians and vehicles (Supplementary Note 1.2). To capture the agent's current behavioral tendency, $r_s^{(t)}$ additionally incorporates the instantaneous speed deviation, $v_\mathrm{s}^{(t)} - v_\mathrm{mean}$, where $v_\mathrm{mean}$ is the average human speed estimated from the full interaction dataset. The SVO angle was then computed as:

$$\theta^{(t)} = \tan^{-1}\left(\frac{r_\mathrm{o}^{(t)}}{r_\mathrm{s}^{(t)}}\right) \tag{3}$$

Because both reward terms are strictly positive, $\theta^{(t)}$ is constrained to the interval [0°, 90°], spanning the continuum from self-interest to altruism. Negative (competitive) SVO values do not arise because the task contains no incentive for benefiting oneself by reducing the other's reward. Smaller values of $\theta^{(t)}$ indicate stronger self-interest, whereas larger values indicate greater altruistic orientation.

**Advantage awareness.** In this study, advantage awareness is operationalized through both situation-dependent and agent-intrinsic asymmetries. The situation-dependent asymmetry is implemented using an initial-state heuristic: given the joint configuration of initial position and speed, agents with a clear initial advantage (e.g., higher speed or a shorter distance to the conflict point) are encouraged to accelerate decisively, whereas disadvantaged agents are encouraged to maintain their speed (Supplementary Note 3.3). The agent-intrinsic asymmetry is implemented using the asymmetric collision penalties (pedestrian: $-150$; vehicle: $-100$) as a decision heuristic. Because the vehicle incurs a smaller collision penalty, it is encouraged to adopt moderately more assertive actions, whereas the pedestrian is encouraged to behave more conservatively. Together, these rules operationalize relative-advantage reasoning in the LLM's decision-making.

## 4.4 LLM distillation and closed-loop evaluation

**LLM policy distillation.** To enable real-time closed-loop deployment, we distilled LLM decision-making into a lightweight student network. The training dataset was constructed from LLM decision rollouts generated during the open-loop evaluation. At each timestep, the network received a 10-dimensional feature vector consisting of the positions, speeds, distances to the conflict point, and time-to-conflict-point for both agents, together with the current timestep and a binary role indicator (0 for pedestrian and 1 for vehicle). All continuous features were normalized. The model operated on a sliding window of the preceding 20 frames (1 s) of observations. The student network consisted of a two-layer LSTM followed by two parallel output heads for intention classification and conditional speed regression. The intention classification head predicted one of three action modes (accelerate, decelerate, or maintain speed), whereas the conditional regression head predicted the target speed conditioned on the inferred intention. The final action was jointly determined by the predicted intention and target speed. The network was trained using a joint loss function: $\mathcal{L} = \mathcal{L}_{\text{intent}} + \lambda\mathcal{L}_{\text{reg}}$, where $\mathcal{L}_{\text{intent}}$ is the cross-entropy loss for intention classification, $\mathcal{L}_{\text{reg}}$ is the mean squared error for target-speed regression, and $\lambda = 1.0$ balances the two terms. Network architecture and training hyperparameters are provided in Supplementary Note 5.

**Human–AI closed-loop evaluation.** To validate the effectiveness of the distilled policies in real-time interaction with humans, we conducted a human–AI closed-loop evaluation. Four student networks distilled from the DeepSeek open-loop data were evaluated, corresponding to the Base, SCoT, RA, and HSN prompting strategies. The experimental task and scenario parameters were identical to those used in the human–human experiment. A total of 24 adults who had not participated in the human–human experiment were recruited, comprising six males under 30 years of age, six females under 30, six males over 30, and six females over 30. Participant details are provided in Supplementary Note 2.2. All participants provided informed consent, and the study procedures were approved by the Institutional Review Board of Tsinghua University. Participants were informed that the experiment followed the same pairing scheme as the human–human experiment, in which two games were conducted simultaneously in each trial. In fact, each participant interacted with a separate distilled agent, resulting in three simultaneous games per trial. This design served two purposes: to minimize behavioral bias arising from participants' awareness of interacting with an AI agent[48], and to improve data collection efficiency. Post-experiment debriefing indicated that no participant suspected a discrepancy between the described and actual pairing scheme.

The experimental factors included agent strategy (four levels: Base, SCoT, RA, or HSN), role (pedestrian or vehicle), initial position (balanced, vehicle-advantaged, or pedestrian-advantaged), and initial speed (balanced, vehicle-advantaged, or pedestrian-advantaged). Because scenario orientation showed only a negligible effect in the human–human experiment (Cramér's $V = 0.067$), it was omitted to simplify the experimental design. The full factorial combination of these factors generated 72 unique experimental conditions (4 × 2 × 3 × 3), all of which were presented once to each participant, resulting in 72 trials completed in six blocks separated by 1-min breaks. Experimental sessions lasted approximately 30 min. Mean self-score and mean total score (the sum of the agent's and its human opponent's scores) were computed for the four agent strategies across all games. For comparison, we also computed the mean self-score and mean total score across all participants in the human–human experiment, providing a population-level human baseline.

**Inter-model pairwise evaluation.** To further evaluate the distilled strategies in direct policy-versus-policy interactions, we conducted pairwise round-robin evaluations among eight distilled agents. In addition to the four strategies distilled from LLMs (Base, SCoT, RA, and HSN), we included four agents distilled from the human interaction dataset: Hum. (trained on all human keyframe decisions), H1–2 (trained only on decisions from C1–C2), H1–4 (trained on C1–C4), and H1–6 (trained on C1–C6). The latter three agents were included to test whether training exclusively on high-quality human data is sufficient to achieve coordination performance beyond the human level. Each pair of strategies played 1,000 complete games under both role assignments (pedestrian and vehicle). To evaluate generalization beyond the experimental conditions, initial positions and speeds were randomly sampled from uniform distributions over their respective ranges rather than being restricted to the nine predefined combinations used in the human interaction experiments (see Supplementary Note 1.3). Mean self-score and mean total score were computed for all eight strategies across all games.

# Data availability

The datasets generated and analyzed during this study, including human experimental trajectory data, open-loop model decision data, and closed-loop interaction evaluation data, are available from the corresponding author upon reasonable request.

# Code availability

The source code for the experimental platform, social norm learning framework, LLM prompting pipeline, and policy distillation used in this study is available from the corresponding author upon reasonable request.

## Acknowledgements

This study is supported by the National Natural Science Foundation of China (52572413).

## Author contributions

Y.Y., S.L., C.L., and B.N. conceived the study. Y.Y. and S.L. designed the methodology, developed the experimental platform, implemented the algorithms, and performed the computational experiments. Y.Y., X.G., and H.S. conducted the human experiments and collected the data. Y.Y. and S.L. analyzed the data and interpreted the results. Y.Y. prepared the figures. Y.Y. and S.L. wrote the original draft of the manuscript. C.L., Q.Z., and B.N. reviewed and revised the manuscript. B.N. supervised the project and acquired funding. All authors discussed the results and approved the final manuscript.

## Competing interests

The authors declare no competing interests.

## Additional information

Supplementary Information is available for this paper.
Correspondence should be addressed to Bingbing Nie.

# Supplementary Information - Learning social norms enhances compatibility in dynamic human-AI coordination

Yi Yang[1,6], Siyuan Liu[1,6], Xin Gao[1], Huamu Sun[1], Chao Liu[3,4,5], Qing Zhou[1], Bingbing Nie[1,2,*]

[1]School of Vehicle and Mobility, Tsinghua University, Beijing, China.
[2]State Key Laboratory of Intelligent Green Vehicle and Mobility, Tsinghua University, Beijing 100084, China.
[3]State Key Laboratory of Cognitive Neuroscience and Learning & IDG/McGovern Institute for Brain Research, Beijing Normal University, Beijing, China.
[4]Beijing Key Laboratory of Safe AI and Superalignment, Beijing, China.
[5]Beijing Institute of AI Safety and Governance, Beijing, China.

[6]These authors contributed equally: Yi Yang and Siyuan Liu.
[*]Correspondence to: nbb@tsinghua.edu.cn

# Contents

## Supplementary Note 1. Experimental platform

This Supplementary Note describes the pedestrian-vehicle dynamic game environment used for human data collection and LLM evaluation. As shown in Fig. S1, the simulation defines the horizontal and vertical axes as the x-axis and y-axis, respectively, with the origin centered geometrically at the intersection. The pedestrian moves along the +x direction, and the vehicle moves along the +y direction.

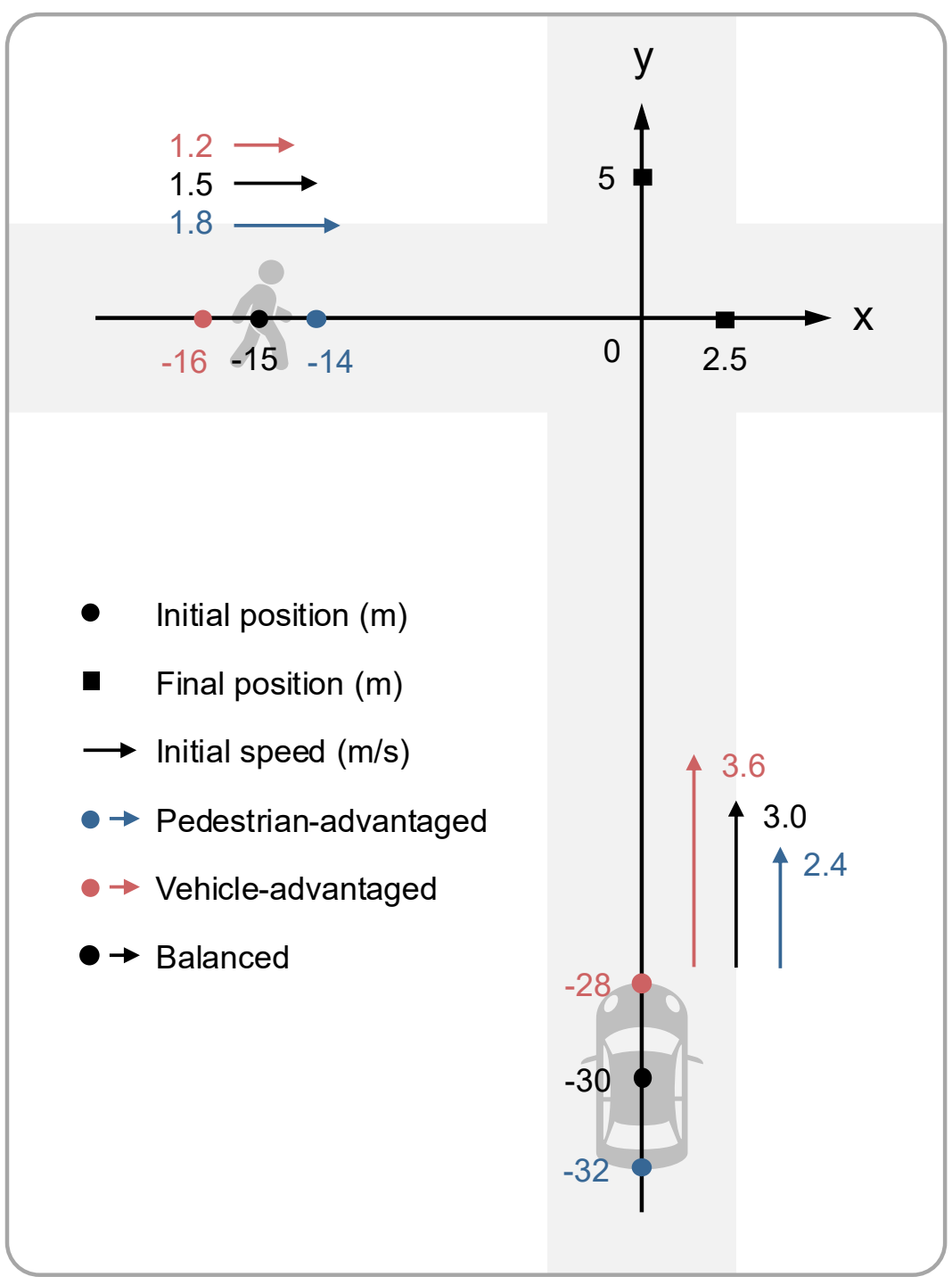

**Fig. S1 | Game environment and initial-condition configurations.** The pedestrian-vehicle game environment is defined on a Cartesian coordinate system centered at the conflict point (origin). The pedestrian moves along the positive x-axis and the vehicle along the positive y-axis. Circles and squares denote the initial and final positions, respectively. Arrows indicate the corresponding initial speeds under the three initial-speed configurations. Black markers indicate the balanced initial condition, whereas blue and red markers represent the pedestrian-advantaged and vehicle-advantaged initial-position configurations. The crossing of the three initial-position and three initial-speed configurations yields nine experimental initial conditions. Detailed spatial layout, kinematic parameters, and initial-condition values are provided in Tables S1–S3, respectively.

### 1.1 Game environment

As shown in Fig. S1, the pedestrian and the vehicle approach a single conflict point along two straight paths. Each agent moves only along its own fixed path, and no lateral deviation is permitted throughout the interaction. Consequently, the interaction is fully described by the two agents' scalar progress along their respective paths. Positions are specified as Cartesian coordinates in meters. The initial positions of the pedestrian and vehicle vary across the experimental conditions (see Section 1.3), whereas their final positions remain fixed. On average, the pedestrian traverses a total path length of 17.5 meters, and the vehicle traverses 35.0 meters before reaching their respective destinations. The two paths meet at the origin, which serves as the potential collision point throughout the experiment. On screen, each agent is rendered as a 2D image, and its effective collision body is defined by the rectangular bounding box of the image. Collision detection is based on whether the two rectangles overlap at any time during the simulation. All coordinate and dimension parameters are detailed in Table S1.

**Table S1.** Spatial layout of the pedestrian and vehicle

| Label | Symbol | Value (m) |
|---|---|---|
| Pedestrian initial position (average) | $P_{0,\text{ped}}$ | (-15.0, 0.0) |
| Pedestrian final position | $P_{*,\text{ped}}$ | (2.5, 0.0) |
| Pedestrian path length (average) | $L_{\text{ped}}$ | 17.5 |
| Vehicle initial position (average) | $P_{0,\text{veh}}$ | (0.0, -30.0) |
| Vehicle final position | $P_{*,\text{veh}}$ | (0.0, 5.0) |
| Vehicle path length (average) | $L_{\text{veh}}$ | 35.0 |
| Conflict point | - | (0.0, 0.0) |
| Effective pedestrian hitbox | $W_{\text{ped}} \times H_{\text{ped}}$ | $1.0 \times 2.0$ |
| Effective vehicle hitbox | $W_{\text{veh}} \times H_{\text{veh}}$ | $2.0 \times 4.0$ |

## 1.2 Kinematics

Both agents move along their fixed paths, and the only controllable variable is speed. At each control step, the active controller (human or agent) issues one of three actions (i.e., accelerate, decelerate, or hold), which are integrated forward using the constant acceleration and deceleration magnitudes given in Table S2. Speed is clamped to the permitted range at every step. As shown in Table S2, all kinematic quantities (speed limits, initial speed, and acceleration and deceleration) of the vehicle are obtained by multiplying the corresponding pedestrian kinematic quantities by a vehicle scale factor $\alpha = 2$. This scaling factor compensates for the fact that the vehicle's average path length (35.0 m) is twice that of the pedestrian (17.5 m), ensuring that under identical behavioral strategies, the two parties arrive at the conflict point at comparable times, thereby keeping the game kinematically fair. For both agents, the deceleration magnitude is twice the acceleration magnitude.

**Table S2.** Kinematic limits and control parameters for the pedestrian and vehicle

| Label | Symbol | Pedestrian | Vehicle |
|---|---|---|---|
| Minimum speed | $v_{\text{min}}$ | 0 m/s | 0 m/s |
| Maximum speed | $v_{\text{max}}$ | 3.0 m/s | 6.0 m/s |
| Initial speed (average) | $v_0$ | 1.5 m/s | 3.0 m/s |
| Acceleration | $a_{\text{acc}}$ | 1.5 m/s$^2$ | 3.0 m/s$^2$ |
| Deceleration | $a_{\text{dec}}$ | 3.0 m/s$^2$ | 6.0 m/s$^2$ |
| Scale factor | $\alpha$ | - | 2 |

Speed updates follow a constant-acceleration integration with control interval $\Delta t$. An acceleration updates speed as $v_{t+1} = \min(v_t + a_{\text{acc}} \cdot \Delta t, v_{\text{max}})$, a deceleration changes speed as $v_{t+1} = \max(v_t - a_{\text{dec}} \cdot \Delta t, v_{\text{min}})$, and a hold action leaves speed unchanged. Position is updated by $p_{t+1} = p_t + v_{t+1} \cdot \Delta t$ along the agent's fixed direction. Once the speed reaches $v_{\text{max}}$ or $v_{\text{min}}$, continuing to accelerate or decelerate will no longer change the speed until an opposite action is issued. The physical simulation, screen rendering, and data logger all operate at the same frequency of 20 Hz, so the control and integration time step $\Delta t$ is 0.05 seconds.

## 1.3 Initial conditions

To introduce controlled asymmetry between the two parties at the start of each trial, initial position and initial speed were each varied across 3 levels (i.e., pedestrian-advantaged, balanced, and vehicle-advantaged), as shown in Fig. S1 and Table S3. In terms of initial position, relative to the balanced configuration, the pedestrian-advantaged condition shifts the pedestrian's initial position forward by 1.0 m and the vehicle's initial position backward by 2.0 m, giving the pedestrian a spatial advantage. The vehicle-advantaged condition does the opposite, shifting the pedestrian backward and the vehicle forward. In terms of initial speed, relative to the balanced configuration, the pedestrian-advantaged condition increases the pedestrian's initial speed by 0.3 m/s and decreases the vehicle's initial speed by 0.6 m/s, giving the pedestrian an advantage in speed. The vehicle-advantaged condition reduces the pedestrian's speed and increases the vehicle's speed. The crossing of initial position and speed configurations yields nine initial conditions, which appear with equal frequency in the experiment.

**Table S3.** Parameters for different initial position and speed configurations

| | Pedestrian-advantaged | Balanced | Vehicle-advantaged |
|---|---|---|---|
| Initial position: Ped / Veh (m) | -14.0 / -32.0 | -15.0 / -30.0 | -16.0 / -28.0 |
| Initial speed: Ped / Veh (m/s) | 1.8 / 2.4 | 1.5 / 3.0 | 1.2 / 3.6 |

In the pairwise round-robin evaluation of distilled strategies, to test generalizability, initial positions and speeds were no longer limited to the nine discrete combinations but were instead sampled at random. Specifically, the pedestrian's initial position offset $P_{\text{off, ped}}$ was randomly sampled from a uniform distribution over $[-1.5, 1.5]$ m, and the vehicle's initial position offset was set as $P_{\text{off, veh}} = -\alpha \cdot P_{\text{off,ped}}$. The pedestrian's initial speed offset $V_{\text{off, ped}}$ was randomly sampled from a uniform distribution over $[-0.45, 0.45]$ m/s, and the vehicle's initial speed offset was set as $V_{\text{off, veh}} = -\alpha \cdot V_{\text{off, ped}}$.

## 1.4 Reward structure

At the beginning of each trial, each player has a fixed initial endowment. Adjusting speed incurs a cost: the initial endowment is proportionally consumed based on the cumulative time the player spends accelerating or decelerating, while maintaining a constant speed consumes no points. Crossing the intersection first yields a bonus. For non-collision trials, a player's score $R$ consists of three components: the initial endowment, the speed change cost, and the first-passage bonus:

$$R = R_{\text{ini}} - C_{\text{change}} + \mathbb{I}(first)R_{\text{first}} \quad \text{(S1)}$$

where $R_{\text{ini}} = 50$ is the initial endowment, $\mathbb{I}(\cdot)$ is an indicator function that equals 1 if the player passes the intersection first and 0 otherwise, $R_{\text{first}} = 50$ is the first-passage bonus, and the speed change cost $C_{\text{change}}$ is a linear function of the cumulative time spent changing speed, with a cap:

$$C_{\text{change}} = \min(k(t_{\text{acc}} + t_{\text{dec}}), C_{\text{max}}) \quad \text{(S2)}$$

where $k = 50/3$ (points per second) is the speed change cost rate, $t_{\text{acc}}$ and $t_{\text{dec}}$ are the cumulative times (in seconds) the player spends accelerating and decelerating during the trial, respectively. $C_{\text{max}} = 50$ is the cap of the speed change cost, which is the same as the initial endowment $R_{\text{ini}}$. Therefore, once the initial endowment is exhausted, further speed adjustments incur no additional point loss.

In the event of a collision, all three score components above are reset to zero and $R$ is replaced by a fixed role-dependent penalty, instead of the normal reward calculation: a collision penalty $P_{\text{veh}} = -100$ for the vehicle and $P_{\text{ped}} = -150$ for the pedestrian. The complete set of reward and penalty parameters is summarized in Table S4.

**Table S4.** Reward and penalty parameters

| Label | Symbol | Value |
|---|---|---|
| Initial endowment | $R_{\mathrm{ini}}$ | 50 |
| First-passage bonus | $R_{\mathrm{first}}$ | 50 |
| Speed change cost rate | $k$ | 50/3 |
| Speed change cost cap | $C_{\max}$ | 50 |
| Vehicle collision penalty | $P_{\mathrm{veh}}$ | -100 |
| Pedestrian collision penalty | $P_{\mathrm{ped}}$ | -150 |

### 1.5 Rule-based opponent

The pairing scheme in the human-human game experiment included a rule-based robot as a fourth player in each group of three participants, enabling two independent games to be conducted simultaneously within each trial. The robot was preset with three strategies, namely aggressive, neutral, and yielding, to increase the diversity of opponent behavior patterns and prevent participants from developing strategic adaptations to specific opponents.

The real-time decision-making of the robot is based on SVO-guided logic with conflict zone triggering. Specifically, the area around the conflict point is defined as the critical decision zone. Under normal circumstances, the robot computes its own SVO in real time and adjusts its speed accordingly to make its SVO converge toward a target value. Once both the robot and the opponent enter the critical decision zone simultaneously, the collision risk increases. To prevent hesitant, oscillatory behaviors from continuous SVO updates that could confuse the opponent, the robot makes a one-time fixed decision to either accelerate and rush through or decelerate and yield, maintaining that decision until the game ends.

The differences among robot types lie in their target SVO values and decision preferences within the critical decision zone. The yielding-type robot has the highest target SVO (60°), tending to decelerate and yield. The neutral-type robot has a moderate target SVO (45°), striking a balance between yielding and passing. The aggressive-type robot has the lowest target SVO (30°), tending to accelerate and rush through. Within the critical decision zone, the robot selects its strategy based on both its inherent preference and the specific situation. In most cases, the yielding-type robot tends to continuously decelerate until coming to a complete stop, and the aggressive-type robot tends to apply maximum acceleration to reach its top speed for a forced passage. Meanwhile, the neutral-type robot flexibly switches between these two boundary strategies based on the real-time dynamic context.

## Supplementary Note 2. Human experiments

This Supplementary Note describes the participants and experimental procedures for the human-human and human-AI interaction studies. All participants are identified solely by anonymized codes; no names or other personally identifiable information are reported. The human-human game experiment included 96 participants, whose codes begin with the letter H, followed by the group and player numbers (e.g., H01-1 denotes the first member of group 1). The human-AI closed-loop validation experiment included 24 participants, whose codes begin with the letter C using the same convention. The two participant samples were non-overlapping. Recruitment for each experiment followed a balanced design with respect to gender and age, forming four equally sized demographic cells: males and females under 30 years, and males and females aged 30 years or older. Within each cell, participants were randomly assigned to fixed groups of three, who participated simultaneously. Table S5 summarizes the demographic composition of the two experimental samples, which were comparable in age.

**Table S5.** Demographic composition of participants in the two experiments

| Characteristic | Human-human experiment | Human-AI experiment |
|---|---|---|
| Participants, *n* | 96 | 24 |
| Groups, *n* | 32 | 8 |
| Sex (M/F), *n* | 48 / 48 | 12 / 12 |
| Age (years), mean ± s.d. | 32.2 ± 10.2 | 32.0 ± 9.9 |
| Age < 30 years (M/F), *n* | 24 / 24 | 6 / 6 |
| Age ≥ 30 years (M/F), *n* | 24 / 24 | 6 / 6 |

## 2.1 Human-human game experiment

In the human-human game experiment, each participant completed a total of 108 interactions, of which 72 were against human opponents and 36 against the rule-based robot opponent. Table S6 lists each participant's ID, gender, age, and scores. The scores include the average score in games against human opponents, the average score in games against the robot opponent, and the overall average score across all games, all rounded to the nearest integer.

**Table S6.** Human-human game experiment participants ($n$ = 96)

| ID | Gender | Age | Scores | | |
|---|---|---|---|---|---|
| | | | Avg. vs. human | Avg. vs. robot | Overall avg. |
| H01-1 | M | 25 | 18 | 35 | 23 |
| H01-2 | M | 28 | 16 | 43 | 25 |
| H01-3 | M | 20 | 15 | 11 | 13 |
| H02-1 | M | 22 | 60 | 40 | 53 |
| H02-2 | M | 19 | 53 | 54 | 54 |
| H02-3 | M | 27 | 56 | 42 | 52 |
| H03-1 | M | 30 | 32 | 25 | 30 |
| H03-2 | M | 28 | 39 | 26 | 35 |
| H03-3 | M | 28 | 50 | 53 | 51 |
| H04-1 | M | 29 | 43 | 25 | 37 |
| H04-2 | M | 26 | 45 | 34 | 41 |
| H04-3 | M | 29 | 33 | 52 | 39 |
| H05-1 | M | 25 | 40 | 43 | 41 |
| H05-2 | M | 25 | 39 | 42 | 40 |
| H05-3 | M | 23 | 47 | 38 | 44 |
| H06-1 | M | 24 | 45 | 51 | 47 |
| H06-2 | M | 24 | 65 | 48 | 60 |
| H06-3 | M | 30 | 61 | 39 | 53 |
| H07-1 | M | 20 | 45 | 34 | 41 |
| H07-2 | M | 24 | 28 | 47 | 34 |
| H07-3 | M | 23 | 39 | 34 | 37 |
| H08-1 | M | 23 | 28 | 18 | 25 |
| H08-2 | M | 19 | 26 | 44 | 32 |
| H08-3 | M | 18 | 12 | 21 | 15 |
| H09-1 | F | 22 | 47 | 31 | 42 |
| H09-2 | F | 23 | 35 | 45 | 38 |
| H09-3 | F | 24 | 34 | 34 | 34 |
| H10-1 | F | 18 | 39 | 32 | 36 |

| ID | Gender | Age | Scores | | |
|---|---|---|---|---|---|
| | | | Avg. vs. human | Avg. vs. robot | Overall avg. |
| H10-2 | F | 21 | 54 | 36 | 48 |
| H10-3 | F | 21 | 58 | 46 | 54 |
| H11-1 | F | 22 | 51 | 25 | 42 |
| H11-2 | F | 28 | 41 | 43 | 41 |
| H11-3 | F | 21 | 46 | 48 | 47 |
| H12-1 | F | 23 | 53 | 29 | 45 |
| H12-2 | F | 27 | 43 | 54 | 47 |
| H12-3 | F | 26 | 46 | 46 | 46 |
| H13-1 | F | 23 | 49 | 35 | 44 |
| H13-2 | F | 29 | 47 | 32 | 42 |
| H13-3 | F | 22 | 40 | 51 | 43 |
| H14-1 | F | 24 | 27 | 19 | 24 |
| H14-2 | F | 19 | 42 | 25 | 36 |
| H14-3 | F | 20 | 21 | 42 | 28 |
| H15-1 | F | 20 | 21 | 8 | 17 |
| H15-2 | F | 27 | 36 | 39 | 37 |
| H15-3 | F | 21 | 31 | 46 | 36 |
| H16-1 | F | 18 | 20 | 38 | 26 |
| H16-2 | F | 24 | 28 | 59 | 38 |
| H16-3 | F | 27 | 23 | 21 | 22 |
| H17-1 | M | 38 | 25 | 27 | 26 |
| H17-2 | M | 38 | 27 | 24 | 26 |
| H17-3 | M | 37 | 26 | 53 | 35 |
| H18-1 | M | 41 | 37 | 49 | 41 |
| H18-2 | M | 32 | 62 | 7 | 44 |
| H18-3 | M | 42 | 42 | 55 | 46 |
| H19-1 | M | 35 | 61 | 58 | 60 |
| H19-2 | M | 31 | 71 | 51 | 64 |
| H19-3 | M | 37 | 60 | 40 | 53 |
| H20-1 | M | 49 | 40 | 26 | 35 |
| H20-2 | M | 55 | 46 | 40 | 44 |
| H20-3 | M | 53 | 31 | 40 | 34 |
| H21-1 | M | 41 | 55 | 35 | 48 |
| H21-2 | M | 38 | 42 | 54 | 46 |
| H21-3 | M | 31 | 53 | 44 | 50 |
| H22-1 | M | 38 | 26 | 48 | 33 |
| H22-2 | M | 33 | 8 | 33 | 16 |
| H22-3 | M | 36 | 5 | 37 | 16 |
| H23-1 | M | 34 | 41 | 22 | 35 |
| H23-2 | M | 44 | 32 | 37 | 34 |
| H23-3 | M | 38 | 36 | 40 | 37 |
| H24-1 | M | 37 | 46 | 52 | 48 |
| H24-2 | M | 48 | 40 | 40 | 40 |
| H24-3 | M | 44 | 54 | 31 | 47 |
| H25-1 | F | 45 | 42 | 23 | 36 |
| H25-2 | F | 41 | 50 | 30 | 42 |

| ID | Gender | Age | Scores | | |
|---|---|---|---|---|---|
| | | | Avg. vs. human | Avg. vs. robot | Overall avg. |
| H25-3 | F | 36 | 43 | 43 | 43 |
| H26-1 | F | 33 | 46 | 52 | 48 |
| H26-2 | F | 39 | 48 | 44 | 47 |
| H26-3 | F | 37 | 58 | 50 | 56 |
| H27-1 | F | 33 | 50 | 30 | 43 |
| H27-2 | F | 33 | 58 | 54 | 57 |
| H27-3 | F | 55 | 49 | 43 | 47 |
| H28-1 | F | 54 | 39 | 31 | 36 |
| H28-2 | F | 49 | 50 | 49 | 50 |
| H28-3 | F | 54 | 49 | 40 | 46 |
| H29-1 | F | 44 | 52 | 42 | 49 |
| H29-2 | F | 32 | 62 | 32 | 52 |
| H29-3 | F | 44 | 56 | 46 | 53 |
| H30-1 | F | 49 | 49 | 51 | 50 |
| H30-2 | F | 30 | 44 | 53 | 47 |
| H30-3 | F | 38 | 60 | 37 | 53 |
| H31-1 | F | 45 | 33 | 36 | 34 |
| H31-2 | F | 46 | 44 | 52 | 46 |
| H31-3 | F | 53 | 35 | 39 | 36 |
| H32-1 | F | 36 | 43 | 33 | 39 |
| H32-2 | F | 39 | 39 | 47 | 42 |
| H32-3 | F | 41 | 50 | 36 | 45 |

## 2.2 Human-AI validation experiment

In the human-AI closed-loop validation experiment, each participant completed a total of 72 interactions with different types of distilled models. Table S7 lists the group membership, gender, age, and the average score across the 72 interactions (rounded to the nearest integer) for each participant.

**Table S7.** Human-AI closed-loop experiment participants (*n* = 24)

| ID | Gender | Age | Average Score |
|---|---|---|---|
| C01-1 | M | 28 | 54 |
| C01-2 | M | 25 | 53 |
| C01-3 | M | 24 | 42 |
| C02-1 | M | 19 | 50 |
| C02-2 | M | 35 | 40 |
| C02-3 | M | 19 | 44 |
| C03-1 | F | 25 | 35 |
| C03-2 | F | 25 | 41 |
| C03-3 | F | 19 | 51 |
| C04-1 | F | 24 | 43 |
| C04-2 | F | 24 | 42 |
| C04-3 | F | 25 | 53 |
| C05-1 | M | 53 | 47 |
| C05-2 | M | 47 | 39 |
| C05-3 | M | 33 | 48 |

| ID | Gender | Age | Average Score |
|---|---|---|---|
| C06-1 | M | 28 | 38 |
| C06-2 | M | 41 | 31 |
| C06-3 | M | 50 | 45 |
| C07-1 | F | 37 | 36 |
| C07-2 | F | 37 | 25 |
| C07-3 | F | 40 | 50 |
| C08-1 | F | 41 | 23 |
| C08-2 | F | 38 | 39 |
| C08-3 | F | 31 | 42 |

# Supplementary Note 3. LLM prompting

This Supplementary Note describes the prompting framework used to evaluate LLMs in the interaction game. Five models were evaluated under an identical prompting protocol: DeepSeek-V4-Flash, Gemini-3.1-Flash-Lite, Doubao-Seed-2.0-Lite, Llama-4-Maverick, and GPT-5.4-Mini. Each model was tested with four prompting strategies (Base, SCoT, RA, and HSN). These strategies shared the same task context and output protocol, differing only in the game history snapshot and the decision instructions. The prompt at each decision step consists of three parts. First, the task context defines the agent's role, environment physics, and scoring mechanism (Section 3.1). Second, a game history snapshot provides up to twenty recent state frames, with the final frame representing the current moment (Section 3.2). Third, a strategy-specific decision task instructs the agent how to reason and what to output (Section 3.3). The agent's only action is to set a target speed for the next second, returned under the strict output protocol in Section 3.4.

## 3.1 Task context

The task context opens with a role definition (Fig. S2), framing the model as a game-theoretic agent at a perpendicular intersection. The agent assumes one role and faces a human opponent in the other, with both adjusting their speeds in real time to cross. Under the RA strategy, the role definition adds a risk-averse personality, where fear of collision outweighs the desire to rush through.

**Role definition**

You are a game-theoretic decision-making agent in a 2D physical environment, playing the role of {self role}. You are at a perpendicular intersection, engaged in real-time gameplay with a human opponent playing the role of {opponent role}. Both parties need to cross the intersection and reach their respective destinations, and can adjust their speeds in real time. For RA: You have a risk-averse personality, where the fear of negative loss (a collision) outweighs the desire for positive gain (successfully crossing).

**Fig. S2 | Role definition prompt.** The first part of this prompt is shared by all strategies, casting the model as a game-theoretic decision-making agent at a perpendicular intersection against a human opponent. The text shown in blue is appended only under the RA strategy.

The prompt then describes the physics rules (Fig. S3). Game time starts at zero and advances at 20 frames per second. Position is a one-dimensional coordinate centered on the intersection: negative values lie before the intersection, positive values beyond it, and distance to the intersection is the absolute value. The initial positions and initial speeds of both agents vary across trials and are read from an external configuration. The estimated time to reach the intersection is distance divided by current speed, capped at 10.

### Physics rules

1. **Time (unit: s)**
   - The system starts timing from 0 (game start) and records data at a frequency of 20 frames per second.
2. **Position (unit: m)**
   - Represented using a one-dimensional coordinate. The center of the intersection is at coordinate 0.
   - Coordinate < 0 means not yet passed the intersection; coordinate > 0 means already passed. The distance to the intersection is the absolute value of the current position coordinate.
   - Pedestrian start coordinate: {ped_initial_x}, destination coordinate: +2.5.
   - Vehicle start coordinate: {veh_initial_y}, destination coordinate: +5.0.
3. **Speed (unit: m/s)**
   - Pedestrian: speed range [0, 3], initial speed: {ped_initial_speed}.
   - Vehicle: speed range [0, 6], initial speed: {veh_initial_speed}.
   - Estimated time to reach the intersection = distance to intersection / current speed, capped at a maximum of 10.
4. **Acceleration (unit: m/s²)**
   - Pedestrian: 1.5 when accelerating, -3.0 when decelerating, 0 when moving at constant speed.
   - Vehicle: 3.0 when accelerating, -6.0 when decelerating, 0 when moving at constant speed.

**Fig. S3 | Physics rules prompt.** This prompt is shared by all strategies, specifying the time, the position coordinate centered on the intersection, the speed ranges and initial speeds, and the fixed acceleration and deceleration.

Finally, as shown in Fig. S4, the prompt defines the scoring mechanism shared by both parties and specifies that the agent's core objective is to maximize this score. The prompt also emphasizes that this is a game of complete information: at any moment, both parties have full knowledge of each other's positions and speeds.

### Core objective

Your core objective is to maximize your own score, which consists of the following three components:

1. **Initial points and speed change cost**
   - Both parties start with +50 initial points.
   - Each acceleration or deceleration action consumes these points. The deduction per action = deduction coefficient × |change in speed|.
     - Pedestrian: acceleration deduction coefficient is 11, deceleration deduction coefficient is 5.5.
     - Vehicle: acceleration deduction coefficient is 22, deceleration deduction coefficient is 11.
   - This portion of points can only be deducted down to 0. Any further speed adjustments after that will incur no additional deduction.
2. **Passage reward**
   - The party that crosses the intersection first and reaches the destination receives +50 points, while the other party receives 0 points.
3. **Collision penalty**
   - At any moment during the game, a collision is determined to have occurred when [pedestrian's distance to intersection < 1.5] and [vehicle's distance to intersection < 3.0].
   - Once a collision occurs, the initial points are reset to zero; the pedestrian receives an additional -150 points, and the vehicle receives an additional -100 points.

The above scoring mechanism applies equally to both parties. This is a game of complete information: you and your human opponent are both aware of each other's positions, speeds, and the scoring mechanism in real time.

**Fig. S4 | Core objective prompt.** This prompt is shared by all strategies and informs the agent that its score consists of three components: initial endowment minus speed change cost, first-passage reward, and asymmetric collision penalty. The agent's objective is to maximize its total score in the game.

## 3.2 Game history snapshot

At each decision step, the prompt appends a snapshot of recent play (Fig. S5). The snapshot lists successive frames, each reporting the position, speed, distance to the intersection, and estimated time to the intersection for both the pedestrian and the vehicle. The final frame is labeled as the current moment. The Base, SCoT, and RA strategies all receive this standard snapshot (Fig. S5, top). The HSN strategy receives an augmented snapshot (Fig. S5, bottom) that extends each frame with two additional types of information. Each party's SVO is appended to its per-frame state. Additionally, each frame carries two interaction-level quantities: the cumulative game entropy and the sum of the two parties' SVO. These additions give the HSN agent access to the signals associated with the three principles required by its decision task (Section 3.3), whereas the other three strategies see only the raw physical state.

For Base / SCoT / RA

**Game history snapshot**

The following is a game history snapshot from the past period of time. The last frame is the current moment.

Frame 1 | Time: {t} | Pedestrian: [Position: {ped_x}, Speed: {ped_speed}, Distance to intersection: {ped_dist}, Estimated time to intersection: {ped_ttcp}] | Vehicle: [Position: {veh_y}, Speed: {veh_speed}, Distance to intersection: {veh_dist}, Estimated time to intersection: {veh_ttcp}]

…

Frame 20 (Current frame) | Time: {t} | Pedestrian: [Position: {ped_x}, Speed: {ped_speed}, Distance to intersection: {ped_dist}, Estimated time to intersection: {ped_ttcp}] | Vehicle: [Position: {veh_y}, Speed: {veh_speed}, Distance to intersection: {veh_dist}, Estimated time to intersection: {veh_ttcp}]

Based on the above game situation, perform the following decision-making tasks:

---

For HSN

**Game history snapshot**

The following is a game history snapshot from the past period of time. The last frame is the current moment.

Frame 1 | Time: {t} | Pedestrian: [Position: {ped_x}, Speed: {ped_speed}, Distance to intersection: {ped_dist}, Estimated time to intersection: {ped_ttcp}, SVO: {ped_SVO}] | Vehicle: [Position: {veh_y}, Speed: {veh_speed}, Distance to intersection: {veh_dist}, Estimated time to intersection: {veh_ttcp}, SVO: {veh_SVO}] | Cumulative entropy: {cum_entropy}, Sum of SVO: {SVO_sum}

…

Frame 20 (Current frame) | Time: {t} | Pedestrian: [Position: {ped_x}, Speed: {ped_speed}, Distance to intersection: {ped_dist}, Estimated time to intersection: {ped_ttcp}, SVO: {ped_SVO}] | Vehicle: [Position: {veh_y}, Speed: {veh_speed}, Distance to intersection: {veh_dist}, Estimated time to intersection: {veh_ttcp}, SVO: {veh_SVO}] | Cumulative entropy: {cum_entropy}, Sum of SVO: {SVO_sum}

Based on the above game situation, perform the following decision-making tasks:

**Fig. S5 | Game history snapshot prompt.** Top: the standard snapshot supplied to the Base, SCoT, and RA strategies, reporting each party's position, speed, distance to the intersection, and estimated time to arrival. Bottom: the augmented snapshot supplied to the HSN strategy, which additionally appends each party's per-frame SVO, along with the per-frame cumulative entropy and sum of SVO. Both snapshot types cover up to twenty most recent frames, with the last frame representing the current moment.

## 3.3 Decision task

The complete text of the decision task prompt is shown in Fig. S6 and Fig. S7. In Base and RA (Fig. S6, top), the LLM is required to output a strategic target speed based on the interaction trends in the snapshot. From the moment of decision, its speed will adjust toward this target under fixed acceleration and hold once the target is reached. The prompt emphasizes that, since both parties can readjust their speeds in subsequent rounds, this target speed should be regarded as an optimal local decision based on current information. It is a short-term, revisable intention rather than an irreversible commitment.

Building upon Base and RA, the SCoT strategy inserts an additional explicit reasoning step before the speed decision (Fig. S6, bottom): the agent must first predict the opponent's motion state over the next second based on the opponent's historical trajectory and current state, and only then set its own strategic target speed accordingly.

For Base / RA

**Decision task: set target speed**

Based on the dynamic interaction trends exhibited by both parties in the above snapshot, set a strategic target speed for yourself for the next 1 second. Starting immediately, your speed will adjust toward this target according to the acceleration settings described above, and will remain constant once the target speed is reached.

Strategic principle: The target speed represents only your short-term intention for the next 1 second. You may still make new adjustments at any time in subsequent rounds; the human opponent may also adjust their intention again at any time. Therefore, please treat this decision as an optimal local trial based on current information, rather than an irreversible long-term instruction.

---

For SCoT

**Decision task**

1. **Predict opponent's intention**
   - Before deciding on your own action, you must first predict the opponent's motion state for the next 1 second, based on characteristics such as the opponent's historical trajectory and current state in the above snapshot.
2. **Set target speed**
   - Based on the above prediction, set a strategic target speed for yourself for the next 1 second. Starting immediately, your speed will adjust toward this target according to the acceleration settings described above, and will remain constant once the target speed is reached.
   - Strategic principle: The target speed represents only your short-term intention for the next 1 second. You may still make new adjustments at any time in subsequent rounds; the human opponent may also adjust their intention again at any time. Therefore, please treat this decision as an optimal local trial based on current information, rather than an irreversible long-term instruction.

**Fig. S6 | Decision tasks for the Base, RA, and SCoT strategies.** Top: the decision task shared by the Base and RA strategies, which asks the agent to set a strategic target speed for the next second directly from the interaction trends in the snapshot. Bottom: the SCoT decision task. This task inserts an explicit reasoning step: the agent must first predict the opponent's motion state for the next second, and only then set its own target speed.

The HSN strategy combines the target-speed decision with a requirement to balance three principles (Fig. S7). First, outcome predictability uses game entropy to measure outcome uncertainty. High-quality interaction requires cumulative entropy below 5.27, while a value above this threshold signals a deadlock that the agent must decisively break. Second, value alignment interprets SVO on a scale from 0° to 90°, where 0° denotes aggressive and 90° denotes yielding. High-quality interaction is characterized by the sum of both parties' SVO angles equaling roughly 90° (one aggressive, one yielding). In the early-to-middle game, a sum near 180° indicates both parties are overly yielding, prompting the agent to take the aggressor role and accelerate. In the middle-to-late game, a sum near 0° indicates both parties are overly aggressive, prompting the agent to take the coordinator role and decelerate. Third, advantage awareness uses 20 frames as a threshold to distinguish the opening from the mid-to-late game. In the opening, the agent accelerates when its initial position and speed confer a relative advantage; otherwise, it maintains its initial speed. In the mid-to-late game, the agent accounts for the asymmetric collision penalties when assessing relative advantage:

vehicles are encouraged to behave more assertively, whereas pedestrians are encouraged to adopt a more conservative and yielding strategy.

For HSN

**Decision task: set target speed**

Based on the dynamic interaction trends exhibited by both parties in the above snapshot, set a strategic target speed for yourself for the next 1 second. Starting immediately, your speed will adjust toward this target according to the acceleration settings described above, and will remain constant once the target speed is reached.

Strategic principle: The target speed represents only your short-term intention for the next 1 second. You may still make new adjustments at any time in subsequent rounds; the human opponent may also adjust their intention again at any time. Therefore, please treat this decision as an optimal local trial based on current information, rather than an irreversible long-term instruction.

**Decision strategy: social norm constraints**

When executing the above speed decision, you must comprehensively balance the following social norms to achieve high-quality gameplay:

1. **Outcome predictability**
   - We use game entropy to quantify the uncertainty of the game outcome. Cumulative game entropy is the time integral of game entropy.
   - In high-quality gameplay, the cumulative game entropy is typically less than 5.27.
   - If the cumulative game entropy exceeds 5.27, it indicates that the game is deadlocked. At this point, you need to take clear actions to break the deadlock.
2. **Value alignment**
   - We use Social Value Orientation (SVO) to quantify the playing styles of both parties. In this task, SVO ranges from 0 to 90 degrees. The closer SVO is to 0 degrees, the more aggressive (hardline) the style; the closer it is to 90 degrees, the more accommodating (yielding) the style.
   - In high-quality gameplay, the sum of both parties' SVO should be close to 90 degrees (one party aggressive, the other accommodating).
   - If, in the early and middle stages of the game, the sum of SVO approaches 180 degrees (both parties too accommodating), resulting in extremely low efficiency, you should act as the aggressor and accelerate through the intersection while ensuring no collision occurs.
   - If, in the middle and late stages of the game, the sum of SVO approaches 0 degrees (both parties too aggressive), posing a collision risk, you should act as the coordinator and actively decelerate.
3. **Advantage awareness**
   - If the input snapshot has fewer than 20 frames, it indicates that the game is in the initial stage. At this point, you should make an appropriate opening decision based on your relative advantage. Considering both parties' initial positions and speeds, you have a certain opening advantage and are advised to accelerate. / you have no clear opening advantage and are advised to maintain your initial speed.
   - If the input snapshot has more than 20 frames, it indicates that the game has entered the middle or late stage. Given the asymmetric collision penalties (pedestrian: -150 points, vehicle: -100 points), when you are the vehicle, you may behave more assertively and maintain priority; when you are the pedestrian, you should adopt a more conservative and yielding strategy.

**Fig. S7 | Decision task for the HSN strategy.** On the basis of the target-speed decision task, this prompt explicitly imposes three social-norm constraints.

## 3.4 Output specification

All four strategies share the strictly enforced output protocol shown in Fig. S8. The agent must return only the numerical value of its target speed, with no surrounding text, prefixes, punctuation, or units. The value must also remain within the role-appropriate speed range, namely [0, 3] m/s for the pedestrian and [0, 6] m/s for the vehicle.

**Output specification**

Directly output the numerical value representing your target speed (e.g., 1.8). It is strictly forbidden to include any additional text descriptions, prefixes, punctuation, unit symbols, etc.
Ensure your choice does not exceed the speed limit (pedestrian range [0, 3], vehicle range [0, 6]).

**Fig. S8 | Output specification.** The strictly enforced output protocol, which requires the agent to return only the numerical value of its target speed, with no additional text, prefixes, punctuation, or unit symbols, and to keep the value within the role-appropriate speed range.

# Supplementary Note 4. Behavioral analysis

This supplementary Note examines how LLM agents act when placed in the exact situations human players encountered. To provide a balanced reference distribution, human decision points were uniformly sampled from the recorded gameplay data, capturing moments where human players explicitly chose to accelerate, decelerate, or maintain constant speed. Under this open-loop replay paradigm, each recorded human decision point was replayed to the LLM to elicit target speeds, which were then mapped into three corresponding action categories: accelerate (Acc), maintain constant speed (Con), or decelerate (Dec). The results are presented as Sankey diagrams (Fig. S9). Each row corresponds to one model (DeepSeek-V4-Flash, Gemini-3.1-Flash-Lite, Doubao-Seed-2.0-Lite, Llama-4-Maverick, and GPT-5.4-Mini), and each column to one prompting strategy (Base, SCoT, RA, and HSN). In each panel, the left nodes represent human decisions and the right nodes represent the corresponding LLM decisions for the same interaction states. Ribbon widths indicate the proportion of decision points mapped from human actions to LLM actions, whereas the percentages beside the right nodes summarize the overall action distribution of the model.

## 4.1 Cross-model behavioral patterns

Four patterns hold consistently across the five models. First, Base and SCoT produce nearly identical action distributions within each model. Adding an explicit opponent motion prediction step (as SCoT does) changes the distribution only marginally, indicating that this form of chain-of-thought reasoning has little effect on action selection in this task. Second, RA consistently shifts behavior toward greater caution and yielding, increasing Dec while reducing Acc. For every model, the RA column has higher Dec and lower Acc than its Base column. Third, each model has a distinct default bias under non-HSN strategies. Without the three principles, DeepSeek-V4-Flash collapses heavily onto Con. Doubao-Seed-2.0-Lite leans toward Acc with very few Dec (a tendency to push ahead). Llama-4-Maverick leans toward Dec with few Con (a tendency to yield). GPT-5.4-Mini is generally conservative, with high Con and Dec exceeding Acc. Gemini-3.1-Flash-Lite is initially the closest to the human profile. Fourth, HSN largely preserves the rationality of human decision-making while correcting some of the errors that humans make. For example, when the game reaches a deadlock, HSN decisively accelerates or decelerates to break the stalemate rather than maintaining a constant speed. HSN aims to approximate high-quality human gameplay, not to simply replicate average human behavior. The HSN columns are closer to the human balanced distribution and exhibit greater symmetry between Acc and Dec than the other three strategies, which reflects this advantage.

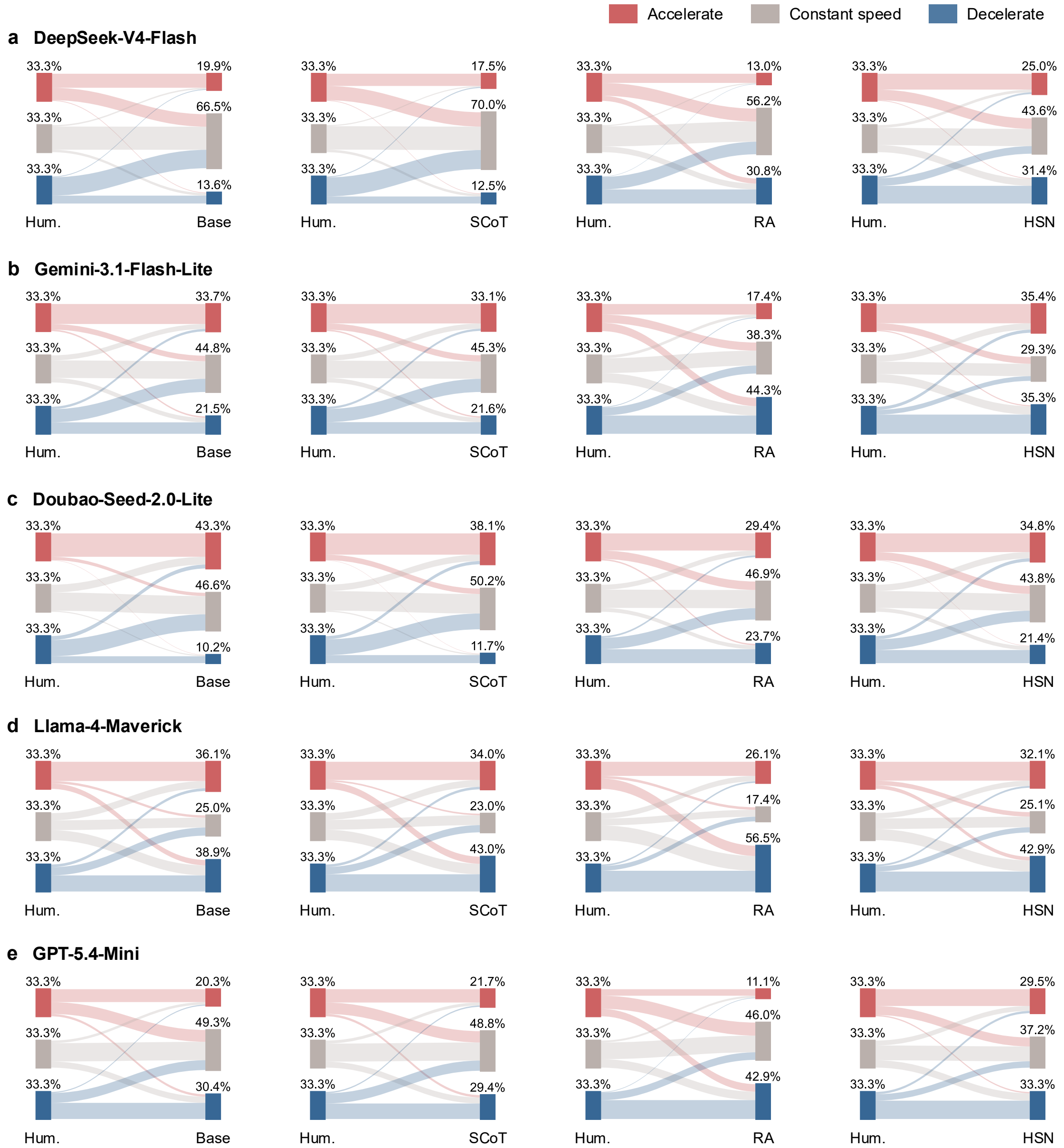

**Fig. S9 | Behavioral correspondence between human and model decisions across five LLMs and four prompting strategies.** Each panel presents a Sankey diagram comparing human decisions (left) with the corresponding model decisions (right) under identical replayed interaction states. Ribbon widths indicate the proportion of decision points mapped from each human action to each model action.

## 4.2 Model-specific behaviors

**DeepSeek-V4-Flash.** Under Base and SCoT, the model collapses strongly onto Con (66.5% and 70.0%), suppressing both Acc and Dec and leaving its decisions far from the human balance. RA raises Dec to 30.8% and lowers Con to 56.2%. HSN brings the distribution closest to the human profile, lowering Con to 43.6% while restoring both Acc (25.0%) and Dec (31.4%).

**Gemini-3.1-Flash-Lite.** Under Base and SCoT, Acc stays near one third (33.7% and 33.1%), with Con only modestly elevated and Dec somewhat reduced. RA shows the strongest risk-averse shift, raising Dec to 44.3% and cutting Acc to 17.4%. HSN yields a nearly symmetric balance (35.4% Acc, 29.3% Con, 35.3% Dec), one of the closest to the human one-third distribution in the entire matrix.

**Doubao-Seed-2.0-Lite.** Under Base and SCoT, the model favors Acc and Con while suppressing Dec to very low levels (10.2% and 11.7%), reflecting a tendency to push ahead and rarely yield. RA raises Dec to 23.7% and lowers Acc to 29.4%, a moderate risk-averse adjustment. HSN sits in between (34.8% Acc, 43.8% Con, 21.4% Dec), with Dec recovering relative to Base but remaining the lowest of the three actions.

**Llama-4-Maverick.** This is the only model biased toward Dec overall. Under Base, Dec is 38.9% while Con is just 25.0%. RA pushes this tendency to an extreme, with Dec reaching 56.5%, the highest in the matrix, and Con compressed to 17.4%. HSN partially curbs this tendency (Dec 42.9%), but the model remains more inclined to Dec than the human reference even under HSN, with Con persistently low.

**GPT-5.4-Mini.** Under Base and SCoT, the model favors Con (around 49%) with Dec exceeding Acc, making it more conservative than the human reference. RA further compresses Acc to 11.1%, among the lowest in the matrix, and raises Dec to 42.9%. HSN again brings the distribution closest to the human profile (29.5% Acc, 37.2% Con, 33.3% Dec), with Dec returning exactly to the one-third human baseline and Acc recovering to near 30%.

# Supplementary Note 5. Policy distillation

To obtain a fast, deployable agent that reproduces the behavior induced by the LLM prompting strategies, we distilled the LLM decision policy into a compact recurrent network. This framework is utilized to distill LLM policies into lightweight agents, and is identically applied to train human-derived models via behavioral cloning (i.e., Hum., H1-2, H1-4, and H1-6 in the main text). The model takes a short window of recent physical state and predicts the agent's intended action for the next step in two coupled forms: a discrete action mode and a continuous target speed. The architecture, illustrated in Fig. S10, consists of a long short-term memory (LSTM) backbone shared by two output heads, an intent-classification head and a conditional residual-regression head. This Supplementary Note documents the model and its training procedure as implemented, and the closed-loop controller that deploys it during live interaction.

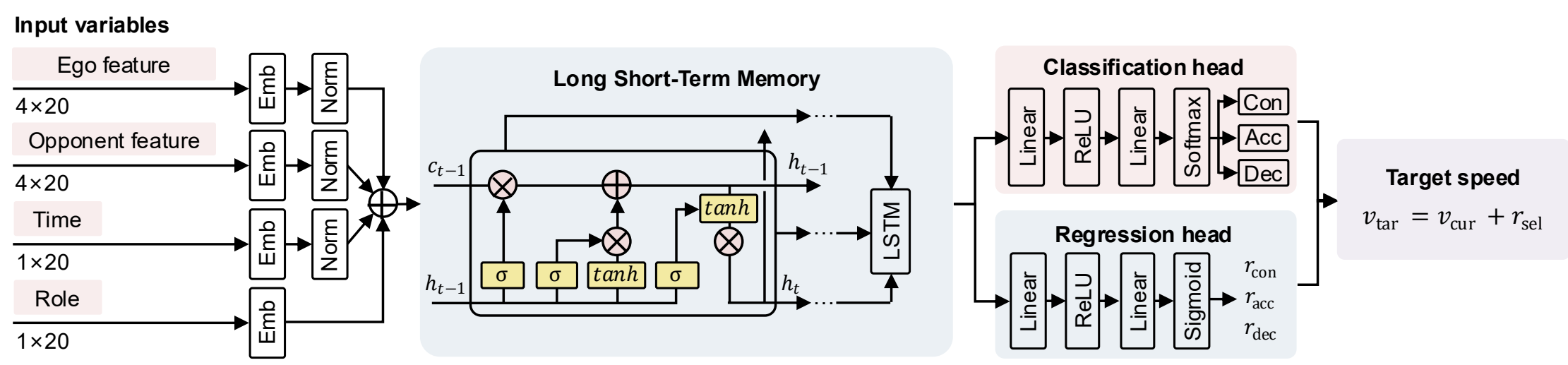

**Fig. S10 | Architecture of the policy network.** A two-layer LSTM backbone feeds into an intent-classification head and a conditional residual-regression head, and the selected residual yields the target speed.

## 5.1 Input representation

Each training example is a sliding window of the twenty most recent frames, and each frame is encoded as a ten-dimensional feature vector. The first four dimensions describe the ego agent and the next four describe the opponent, in each case as position, speed, distance to the intersection, and estimated time to the intersection. The ninth dimension is the elapsed game time, and the tenth is a role identifier set to one when the ego agent is the vehicle and zero when it is the pedestrian. Distance and estimated time to the intersection are derived from raw position and speed in the same way as during live play. The nine physical features are normalized to the unit range with a min-max scaler fitted on the full set of training samples, and the fitted scaler is saved alongside the model so that identical normalization can be applied at inference time. The role identifier is appended after normalization and is not scaled. The same scene is encoded from the ego agent's perspective by swapping the ego and opponent feature blocks, so that a single model serves both the pedestrian and vehicle roles. When a decision point occurs earlier than the twentieth frame of a trajectory, the window is left-padded by repeating its first frame so that every input has a fixed length of twenty.

## 5.2 Network architecture

The network architecture is illustrated in Fig. S10. The backbone consists of a two-layer Long Short-Term Memory (LSTM) network with an input size of 10 and a hidden size of 128. A dropout rate of 0.2 is applied between the two LSTM layers, and the input sequence of 20 frames is processed in temporal order. The hidden state of the final LSTM layer at the last time step is fed into two parallel prediction heads. The intent-classification head consists of a fully connected layer with 64 units, followed by a ReLU activation and a second linear layer that outputs three logits, one for each action mode: maintain constant speed (Con), accelerate (Acc), and decelerate (Dec). During inference, the logits are converted into action probabilities using a softmax function, and the action mode with the highest probability is selected. The conditional residual-regression head uses the same fully connected-ReLU-fully connected architecture and produces three outputs. Each output represents the speed residual, defined as the difference between the target speed and the current speed, for one of the three action modes. The three residuals are denoted by $r_{\text{acc}}$, $r_{\text{con}}$, and $r_{\text{dec}}$, respectively. During inference, the predicted action mode determines which residual is selected, denoted by $r_{\text{sel}}$. The target speed $v_{\text{tar}}$ is then computed by adding the selected speed residual to the current speed $v_{\text{cur}}$.

## 5.3 Training targets

Each labeled decision point provides the agent's current speed and target speed, from which two supervision targets are constructed: an action label and a speed residual. The speed residual is defined as the difference between the target speed and the current speed. The action label is obtained by thresholding the speed residual after normalization by the role-specific maximum speed. Residuals with an absolute normalized value below 0.05 are labeled as maintain constant speed (Con), positive residuals above the threshold as accelerate (Acc), and negative residuals below the threshold as decelerate (Dec). The same target-construction procedure is applied to both LLM-derived and human-derived training samples.

## 5.4 Optimization

The model is trained with a combined objective that sums an intent-classification loss and a residual-regression loss with equal weights. The classification term is the cross-entropy between the predicted logits and the action label. The regression term is the mean squared error between the regression output of the head branch corresponding to the ground-truth action and the normalized residual target, so that only the branch matching the labeled mode is supervised on each example. Optimization uses Adam with a learning rate of 0.001 and a batch size of 256, trained for 100 epochs. All random seeds are fixed at 42 and deterministic backend settings are enabled, so that training runs are reproducible. The fitted feature scaler and the trained model weights are saved to disk on completion for use by the deployed controller.

## 5.5 Closed-loop controller

In closed-loop interaction, the distilled model is wrapped in a controller that converts its target-speed output into executable actions (accelerate, decelerate, or maintain constant speed). The model is queried every 0.5 s and predicts a candidate target speed. Between successive queries, the controller uses a deadband rule to drive the agent toward the current target speed: it accelerates or decelerates when the speed deviation exceeds the deadband and otherwise maintains the current speed. A newly predicted target speed is adopted only if the previous target has been substantially reached and the recent state has remained stable. Otherwise, the controller continues tracking the previous target. This design enables the model to update high-level speed intentions every 0.5 s while the deadband controller generates smooth, stable low-level actions at the full simulation rate.

# Supplementary Note 6. Outcome prediction model

The principle of outcome predictability requires a quantitative measure of game-outcome uncertainty. We obtain this measure from an outcome prediction model, a recurrent network that reads a short window of recent interaction states and predicts a probability distribution over these three terminal outcomes: the pedestrian passes first, the vehicle passes first, or the two collide. The entropy of this predicted distribution is the game entropy used throughout the study, and its time integral is the cumulative entropy encoded in the HSN prompt. The network architecture is shown in Fig. S11.

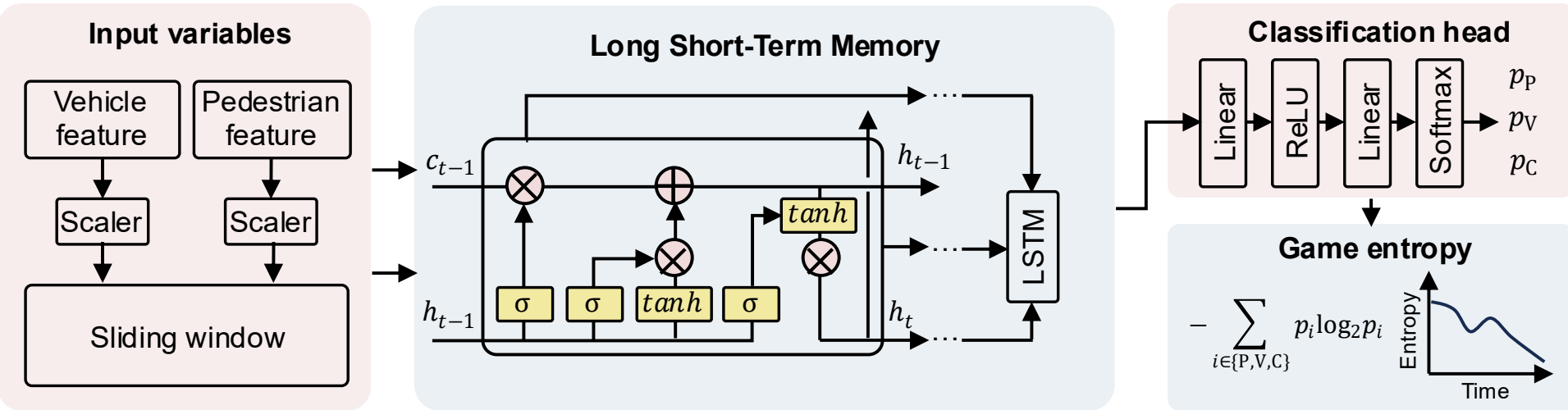

**Fig. S11 | Architecture of the outcome prediction model.** A two-layer LSTM (hidden size 128) encodes a twenty-frame window of eight physical features and classifies the game outcome into three classes. A temperature-scaled softmax (T = 4) yields the outcome probabilities, whose Shannon entropy defines the game entropy.

## 6.1 Input representation

Each input is a sliding window of the twenty most recent frames, and each frame is encoded as an eight-dimensional feature vector. Four dimensions represent the pedestrian and four represent the vehicle, in each case as position, speed, distance to the intersection, and estimated time to the intersection. Distance is the absolute value of the position coordinate, and estimated time to the intersection is the distance divided by the current speed, clipped to the range of zero to ten. The eight features are normalized with a scaler fitted on the full set of training games, and the fitted scaler is saved so that identical normalization can be applied at inference time. When a frame occurs within the first twenty frames of a game, its window is left-padded by repeating the first available frame, so that every input has a fixed length of twenty.

## 6.2 Network architecture

The backbone is a two-layer LSTM with an input size of eight and a hidden size of 128, using a dropout rate of 0.2 between layers and processing the twenty-frame window in temporal order. The hidden state of the final layer at the last time step is passed to a classification head consisting of a linear layer of 64 units, a ReLU nonlinearity, a dropout of 0.1, and a final linear layer producing three logits. A softmax layer then produces the outcome probabilities ($p_{\mathrm{P}}, p_{\mathrm{V}}, p_{\mathrm{C}}$), corresponding to pedestrian-first, vehicle-first, and collision outcomes, respectively. The architecture corresponds to the diagram in Fig. S11.

## 6.3 Training

Each game is labeled by its recorded terminal outcome, read from the final summary row of its log and mapped to one of the three classes. Every frame within a game inherits that game's outcome label, so the network learns to predict the eventual outcome from any window of partial play, including early windows where the outcome is still genuinely uncertain. Distance and estimated-time features are derived in the same way as during live play, so that the training and inference representations match. The model is trained with the cross-entropy loss over the three outcome classes, with equal class weights. Optimization uses Adam with a learning rate of 0.001 and a batch size of 256, and the model is trained for 25 epochs. The trained weights and the fitted feature scaler are saved for entropy computation during deployment..

### 6.4 Entropy computation

At inference time, the outcome prediction model is applied frame by frame to produce a probability distribution over the three outcomes. The logits are divided by a fixed temperature before the softmax to smooth short-term fluctuations in the predicted probabilities. The game entropy at each frame is the Shannon entropy, in bits, of the resulting three-class distribution, and the cumulative game entropy is its temporal integral over the course of the game. These quantities are supplied to the HSN agent through the augmented snapshot and are used to define the predictability threshold in the HSN decision task.

## Supplementary Note 7. Reinforcement learning validation of social norms

This Supplementary Note presents a reinforcement learning (RL) implementation of the three principles underlying the human social norms introduced in the main text. To examine whether these principles generalize beyond prompt-based reasoning, two RL agents were trained in the same pedestrian-vehicle interaction game. RL-Base optimized only the task reward, whereas RL-HSN incorporated additional reward terms that explicitly encouraged outcome predictability, value alignment, and advantage awareness. Their performance was compared with that of their LLM counterparts to evaluate whether the proposed social norms generalize across AI architectures.

The experimental results demonstrate that RL-HSN achieves performance comparable to LLM-HSN, whereas RL-Base fails to achieve comparable performance. Moreover, RL-HSN exhibits behavioral characteristics consistent with successful human coordination, including lower game entropy, complementary SVOs, and role-appropriate asymmetry. In contrast, the poor performance of RL-Base indicates that these behaviors do not emerge from task optimization alone. Together, these findings suggest that the proposed principles provide an architecture-independent representation of social norms that benefits both reasoning-based and reinforcement learning agents.

### 7.1 Experimental design

To evaluate whether the three principles generalize from LLM prompting to reinforcement learning, we trained two RL agents in the pedestrian-vehicle dynamic game environment of this study. The two agents, denoted RL-Base and RL-HSN, share identical state representation, action space, network architecture, opponent randomization scheme, and Proximal Policy Optimization (PPO) training procedure. They differ only in their reward functions: RL-Base optimizes only the task score, whereas RL-HSN augments the task score with explicit rewards for adherence to the three principles (outcome predictability, value alignment, and advantage awareness). Comparing these two agents isolates the effect of social-norm encoding from other confounding factors.

**State and action spaces.** The interaction is modeled as a Markov decision process $M = (\mathcal{S}, \mathcal{A}, \mathcal{P}, r, \gamma)$. The state $s_t \in \mathcal{S}$ consists of a window of the $K = 20$ most recent frames. Let $x_t$ denote the input feature vector at time $t$ as defined in Section 5.1. Then the state is:

$$s_t = (x_{t-K+1}, \dots, x_t) \tag{S3}$$

The action space is discrete:

$$\mathcal{A} = \{\text{con}, \text{acc}, \text{dec}\} \tag{S4}$$

corresponding to maintaining speed, accelerating, and decelerating, respectively. The reward function $r(s_t, a_t, s_{t+1})$ differs between RL-Base and RL-HSN and is specified in detail in Section 7.2.

**Network architecture.** The network architecture follows the LSTM-based design described in Section 5.2. The policy is initialized with the weights of the human-distilled model and uses a two-layer LSTM (hidden size 128) to encode the historical state window. For PPO training, the model is extended with an actor-critic structure:

$$\pi_\theta(a_t|s_t) = \text{Softmax}\big(f_\theta^\pi(h_t)\big), \qquad V_\phi(s_t) = f_\phi^V(h_t) \tag{S5}$$

where $h_t = \text{LSTM}(s_t) \in \mathbb{R}^{128}$ is the LSTM hidden state. The value head is a two-layer MLP with layer sizes of 128, 64, and 1 and ReLU activation. After training, only the policy component is retained for closed-loop deployment.

**PPO optimization.** The agent is trained using PPO. Generalized Advantage Estimation (GAE) is used for advantage computation. The policy loss employs the standard PPO clipping objective:

$$L_\pi(\theta) = -E_t[\min(q_t(\theta)\hat{A}_t, \text{clip}(q_t(\theta), 1-\epsilon, 1+\epsilon)\hat{A}_t)] \tag{S6}$$

where $q_t(\theta) = \pi_\theta(a_t|s_t)/\pi_{\theta_{\text{old}}}(a_t|s_t)$ is the probability ratio. The value loss is the mean squared error between predicted returns and target returns, and the total loss includes an entropy bonus to encourage exploration. All hyperparameters are reported in Table S8.

**Table S8.** PPO hyperparameters

| Parameter | Value |
|---|---|
| Learning rate | $3 \times 10^{-5}$ |
| Discount factor $\gamma$ | 0.99 |
| GAE parameter $\lambda$ | 0.95 |
| Clip range $\varepsilon$ | 0.2 |
| Value loss coefficient $c_v$ | 0.5 |
| Entropy coefficient $c_h$ | 0.01 |
| Number of parallel environments | 1 |
| Steps per update | 256 |
| Epochs per update | 4 |
| Mini-batch size | 256 |
| Gradient norm clip | 0.5 |

**Opponent randomization.** During training, an opponent policy is randomly sampled at the beginning of each episode from a pool $\mathcal{P}_{\text{opp}}$ of frozen human-distilled and LLM-distilled models. This randomization exposes the RL agent to diverse interaction styles, preventing overfitting to a single opponent behavior and improving generalization to unseen partners. Each RL agent was trained for 256,000 decision steps.

## 7.2 Reward functions

The total reward consists of a process reward and a terminal reward:

$$r_t = r_t^{\text{proc}} + \mathbb{I}_{\text{terminal}}(t) r^{\text{term}} \tag{S7}$$

Let the agent's progress, action cost, and switching cost be defined as follows:

$$\Delta p_t = \max(0, p_{t+1}^{\text{ego}} - p_t^{\text{ego}}),\ \ c_t^{\text{act}} = \mathbb{I}(a_t \neq \text{con}),\ \ c_t^{\text{sw}} = \mathbb{I}(a_t \neq a_{t-1}) \tag{S8}$$

To mitigate the sparsity of terminal rewards, we use we use the pre-trained outcome prediction model to estimate the probabilities of pedestrian-first, vehicle-first, and collision (see Supplementary Note 6):

$$(P_t^{\text{ped}}, P_t^{veh}, P_t^{\text{col}}) = g_\psi(s_t) \tag{S9}$$

The agent's expected utility $U_t^{\text{ego}}$ is then defined accordingly. For a pedestrian agent:

$$U_t^{\text{ego}} = 100P_t^{\text{ped}} + 50P_t^{\text{veh}} - 150P_t^{\text{col}} \tag{S10}$$

For a vehicle agent:

$$U_t^{\text{ego}} = 100P_t^{\text{veh}} + 50P_t^{\text{ped}} - 100P_t^{\text{col}} \tag{S11}$$

The utility increment is $\Delta U_t^{\text{ego}} = U_{t+1}^{\text{ego}} - U_t^{\text{ego}}$.

**7.2.1 RL-Base**

The process reward for RL-Base is:

$$r_t^{\text{Base}} = \alpha_p \Delta p_t - \alpha_a c_t^{\text{act}} - \alpha_s c_t^{\text{sw}} + \alpha_u \frac{\Delta U_t^{\text{ego}}}{100} - \alpha_c P_t^{\text{col}} \tag{S12}$$

The terminal reward is defined as the scaled actual ego score:

$$r_{\text{Base}}^{\text{term}} = \beta R^{\text{ego}} \tag{S13}$$

**7.2.2 RL-HSN**

RL-HSN extends RL-Base by adding social-norm reward terms. Define the total welfare of both parties as:

$$W_t = U_t^{\text{ego}} + U_t^{\text{opp}},\ \ \Delta W_t = W_{t+1} - W_t \tag{S14}$$

Let the cumulative game entropy be $E_t$ and the sum of SVO angles be $S_t$. The social-norm penalty terms are defined as:

$$C_t^{\text{ent}} = \max(0, E_t - E_0), \quad C_t^{\text{SVO}} = \frac{|S_t - 90|}{90} \tag{S15}$$

Here, $E_0$ is the entropy threshold for high-uncertainty interactions, and $S_t \approx 90°$ indicates complementary SVOs between the two parties.

The process reward for RL-HSN is:

$$r_t^{\text{HSN}} = r_t^{\text{Base}} + \alpha_w \frac{\Delta W_t}{100} - \alpha_e C_t^{\text{ent}} - \alpha_{\text{SVO}} C_t^{\text{SVO}} \tag{S16}$$

The terminal reward remains the same:

$$r_{\text{HSN}}^{\text{term}} = \beta R^{\text{ego}} \tag{S17}$$

Thus, RL-HSN not only optimizes its own score but also encourages low collision risk, low interaction uncertainty, complementary SVOs, and higher total welfare for both parties.

### 7.2.3 Reward coefficients

All reward coefficients were tuned via grid search on a validation set comprising 5% of human interaction data. The final values are summarized in Table S9.

**Table S9.** Reward coefficients for RL-Base and RL-HSN

| Coefficient | Parameter | Base value | HSN value |
|---|---|---|---|
| $\beta$ | Terminal reward scaling factor | 0.020 | 0.020 |
| $\alpha_p$ | Progress weight | 0.035 | 0.025 |
| $\alpha_a$ | Action cost weight | 0.003 | 0.003 |
| $\alpha_s$ | Switch cost weight | 0.002 | 0.003 |
| $\alpha_u$ | Expected utility weight | 0.010 | 0.008 |
| $\alpha_c$ | Collision probability weight | 0.015 | 0.030 |
| $\alpha_w$ | Social welfare weight | - | 0.010 |
| $\alpha_e$ | Entropy penalty weight | - | 0.010 |
| $\alpha_{\text{SVO}}$ | SVO penalty weight | - | 0.020 |
| $E_0$ | Entropy threshold | - | 5.27 |

## 7.3 Behavioral evaluation

### 7.3.1 Training dynamics

Fig. S12 presents the training dynamics of RL-Base and RL-HSN over the course of PPO updates, tracking three behavioral metrics: mean collision probability, mean sum of SVO, and mean cumulative entropy. These metrics are scale-invariant and task-relevant, enabling direct comparison between the two agents despite their different reward functions.

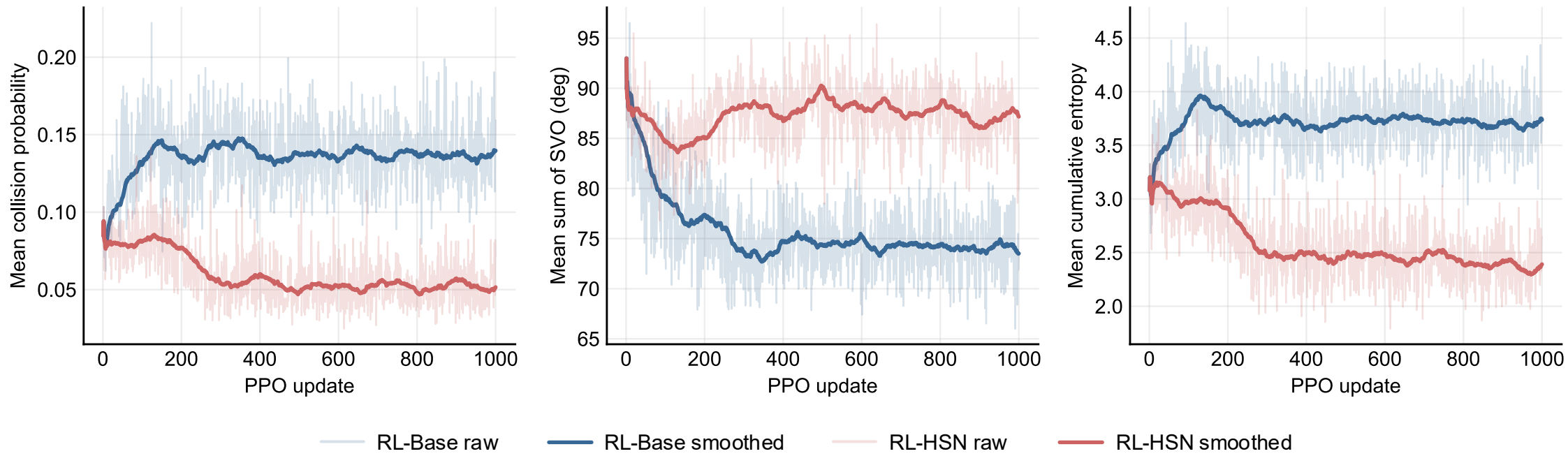

**Fig. S12 | Training dynamics of RL-Base and RL-HSN over the course of PPO updates.** Curves are smoothed using a moving average filter with a window size of 50.

**Collision probability.** RL-Base maintains consistently high collision probability throughout training, eventually stabilizing at around 0.15. In contrast, RL-HSN's collision probability steadily decreases and converges to approximately 0.05. This indicates that explicit social-norm rewards can effectively promote safer behavior, whereas task-only optimization fails to mitigate collision risk.

**Sum of SVO.** As training progresses, RL-Base's sum of SVO converges to 75° and remains stable, indicating that its behavioral pattern is biased toward aggression and fails to establish complementary SVOs with the opponent. In contrast, RL-HSN's SVO sum stabilizes around 90°, an angle that corresponds to the complementary SVO configuration required for successful coordination.

**Cumulative entropy.** RL-Base maintains a high level of cumulative entropy training, reflecting persistent outcome uncertainty. In contrast, RL-HSN's cumulative entropy gradually decreases as training progresses and converges to approximately 2.5. This indicates that interactions involving RL-HSN are more predictable and less uncertain, two key features of successful coordination.

Across all three metrics, RL-HSN consistently outperforms RL-Base, converging to lower collision probability, a sum of SVO angles close to the complementary 90° configuration, and lower cumulative entropy. RL-Base shows no systematic improvement in any metric, confirming that task-only optimization is insufficient to induce socially coordinated behavior.

### 7.3.2 Pairwise evaluation

To evaluate how RL-Base and RL-HSN perform in direct policy-versus-policy interactions and to compare them with the LLM-distilled and imitation-learning strategies reported in the main text, we extended the pairwise round-robin tournament to include the two RL agents. The complete set of ten policies comprised four LLM-distilled strategies (LLM-Base, LLM-SCoT, LLM-RA, and LLM-HSN, all distilled from DeepSeek-V4-Flash), four imitation-learning strategies (Hum. trained on all human keyframe decisions, H1–6 trained on C1–C6, H1–4 trained on C1–C4, and H1–2 trained on C1–C2), and two RL-trained strategies (RL-Base and RL-HSN). Each pair of strategies played 1,000 complete games under both role assignments, with initial positions and speeds randomly sampled from uniform distributions within their respective ranges (see Section 1.3). The evaluation metrics were the mean self-score and mean total score for each strategy, computed across all games in which it participated.

**Overall performance ranking.** Fig. S13 shows the mean self-score and mean total score for each strategy across all pairwise interactions. Among all ten policies, RL-HSN achieves the highest self-score, followed closely by LLM-HSN. RL-HSN also ranks first in total score, indicating that its superior performance does not come at the expense of its opponent but instead reflects mutually beneficial coordination. RL-Base also achieves a relatively high self-score but a notably lower total score. This suggests that its strong individual performance comes primarily at the expense of its opponent rather than through mutually beneficial coordination. The overall performance of the remaining strategies is consistent with the results observed in the main text.

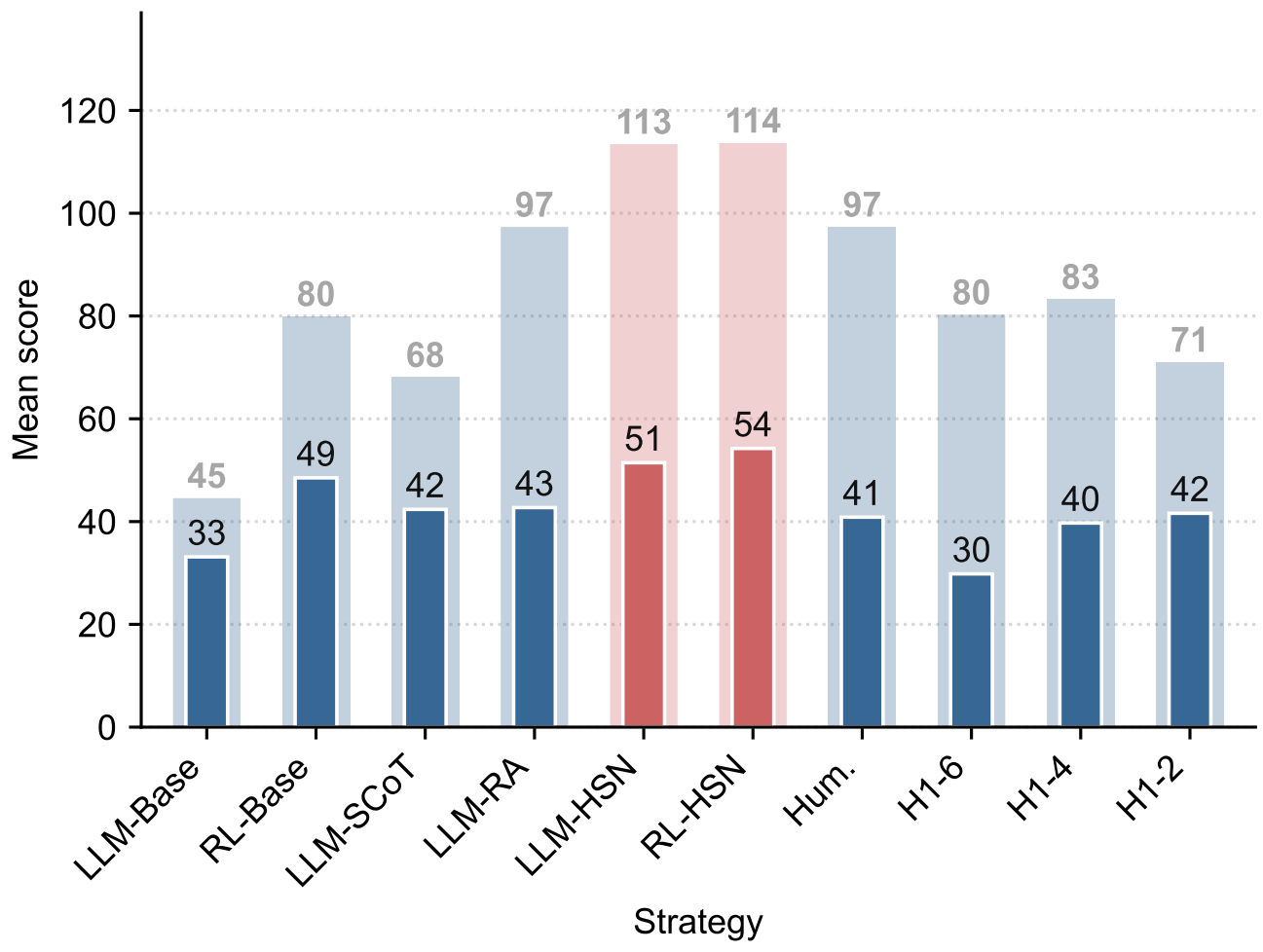

**Fig. S13 | Mean self-scores (dark bars) and mean total scores (light bars) of different strategies in pairwise closed-loop games.**

**Pairwise score matrix.** Fig. S14 shows the mean pedestrian and vehicle scores for each strategy pair. RL-HSN and LLM-HSN consistently maintain high and well-balanced scores for both players across all strategy pairings. Even when facing aggressive opponents, they effectively reduce collision risk by yielding proactively when needed while maintaining high self-scores. In contrast, RL-Base performs poorly against aggressive opponents. In particular, when paired as the vehicle against LLM-Base, both players' scores approach zero, revealing its lack of proactive risk avoidance.

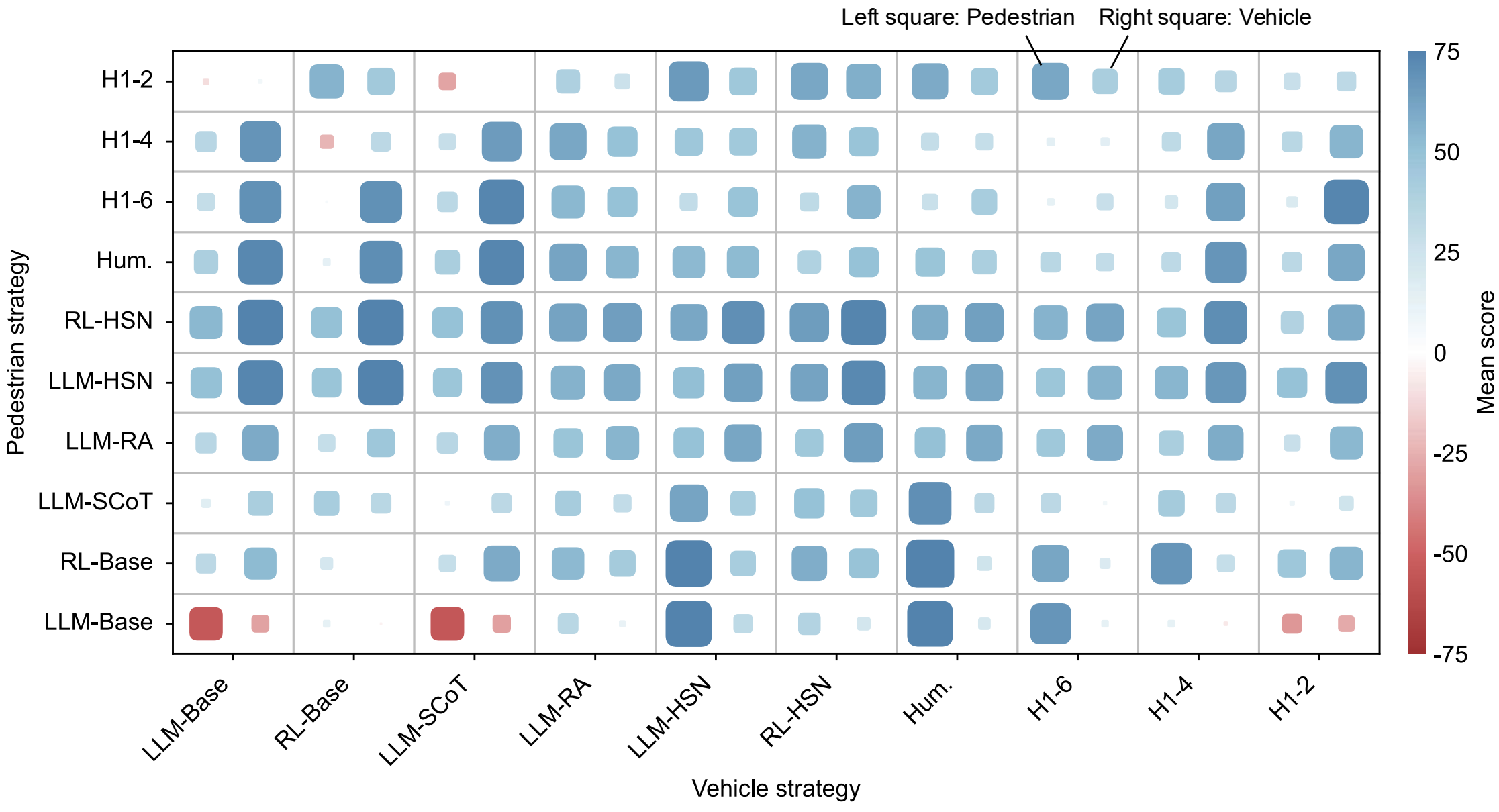

**Fig. S14 | Mean joint score matrix for pairwise closed-loop interactions.** Each cell shows the pedestrian score (left square) and vehicle score (right square). Color indicates the mean score, and square size is proportional to its absolute value.

The pairwise evaluation confirms that RL-HSN achieves coordination performance comparable to LLM-HSN, the best among the LLM-distilled strategies, whereas RL-Base fails to achieve comparable performance. These results reinforce the finding in the main text that explicit encoding of social norms, whether as prompts or as reward terms, produces effective and mutually beneficial coordination, whereas task-only optimization fails to induce socially compatible behavior.